\documentclass[times,twocolumn,final,authoryear]{elsarticle}

\usepackage{ycviu}
\usepackage{framed,multirow}

\usepackage{amssymb}
\usepackage{latexsym}

\usepackage{url}
\definecolor{newcolor}{rgb}{.8,.349,.1}

\usepackage{hyperref}
\usepackage{float}
\usepackage{graphicx}
\usepackage{comment}
\usepackage{amsmath} 
\usepackage{color}
\usepackage[table]{xcolor}
\usepackage{tabularx,ragged2e}
\usepackage{booktabs}
\usepackage{subcaption}
\usepackage{hyphenat}
\usepackage{pifont}
\usepackage{soul}
\newcommand{\cmark}{\ding{51}}%
\newcommand{\xmark}{\ding{55}}%

\journal{Computer Vision and Image Understanding}

\newcommand{\modelname}{SpatioTemporal Alignment with Language}
\newcommand{\modelnameabbr}{STAL}

\newcommand{\ie}{\textit{i.e.}}
\newcommand{\eg}{\textit{e.g.}}
\newcommand{\cf}{\textit{cf.}}
\newcommand{\etal}{\textit{et al.}}
\newcommand{\wrt}{\textit{w.r.t.}}

\renewcommand{\paragraph}[1]{\noindent{\bf #1}}

\begin{document}

\begin{frontmatter}

\title{Finding Moments in Video Collections Using Natural Language}

\author[1]{Victor  \snm{Escorcia}\corref{cor1}} 
\author[1]{Mattia  \snm{Soldan}\corref{cor1}} 
\author[2]{Josef \snm{Sivic  }} 
\author[1]{Bernard \snm{Ghanem }} 
\author[3]{Bryan   \snm{Russell}}

\cortext[cor1]{Authors contributed equally.}

\address[1]{King Abdullah University of Science and Technology (KAUST), Saudi Arabia}
\address[2]{Czech Institute of Informatics, Robotics, and Cybernetics, Czech Technical University in Prague}
\address[3]{Adobe Research}

\received{1 May 2013}
\finalform{10 May 2013}
\accepted{13 May 2013}
\availableonline{15 May 2013}
\communicated{S. Sarkar}

\begin{abstract}
We introduce the task of retrieving relevant video moments from a large corpus of untrimmed, unsegmented videos given a natural language query. 
Our task poses unique challenges as a system must efficiently identify both the relevant videos and localize the relevant moments in the videos. 
To address these challenges, we propose SpatioTemporal Alignment with Language (STAL), a model that represents a video moment as a set of regions within a series of short video clips and aligns a natural language query to the moment's regions. 
Our alignment cost compares variable-length language and video features using symmetric squared Chamfer distance, which allows for efficient indexing and retrieval of the video moments. 
Moreover, aligning language features to regions within a video moment allows for finer alignment compared to methods that extract only an aggregate feature from the entire video moment.
We evaluate our approach on two recently proposed datasets for temporal localization of moments in video with natural language (DiDeMo~\cite{hendricks2017localizing} and Charades-STA~\cite{gao2017tall}) extended to our video corpus moment retrieval setting. 
We show that our \modelnameabbr{} re-ranking model outperforms the recently proposed Moment Context Network~\cite{hendricks2017localizing} on all criteria across all datasets on our proposed task, obtaining relative gains of 37\%-118\% for average recall and up to 30\% for median rank.
Moreover, our approach achieves more than $130\times$ faster retrieval and $8\times$ smaller index size with a 1M video corpus in an approximate setting.
\end{abstract}


\end{frontmatter}

\section{Introduction}

\begin{figure*}[t]
	\centering
	\begin{subfigure}[b]{0.65\linewidth}
    \includegraphics[width=\linewidth]{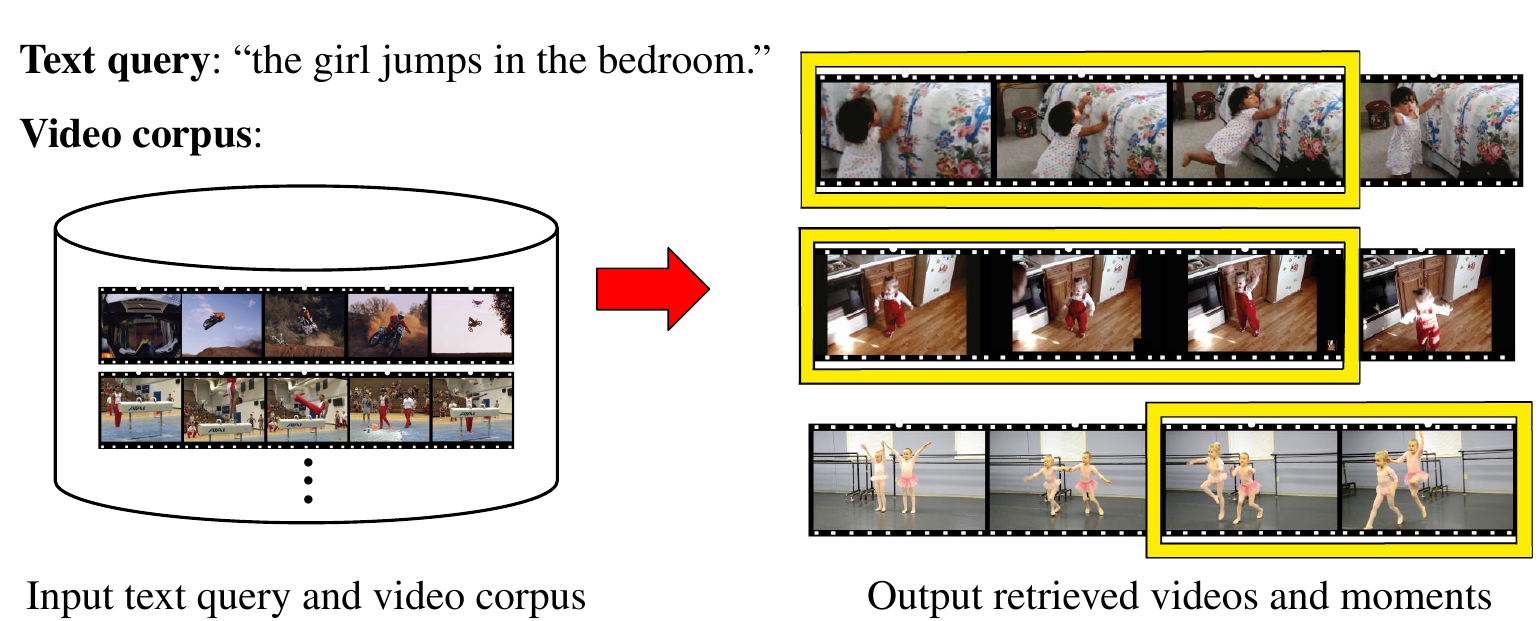}
    \caption{Problem statement \label{fig:problem_statement}}
    \end{subfigure}
    \vrule
    \begin{subfigure}[b]{0.34\linewidth}
    \includegraphics[width=\linewidth]{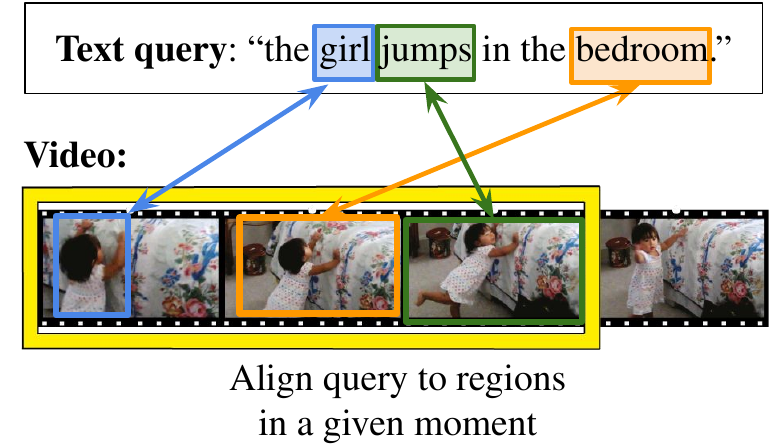}
	\vspace{0.01\linewidth}
    \caption{Approach overview \label{fig:overview}}
    \end{subfigure}
	\caption{
	{\bf \footnotesize Problem statement and approach overview.} \footnotesize (a) Given a natural language query, we seek to find relevant videos from a large corpus of untrimmed, unsegmented videos and temporally localize relevant moments within the returned videos. 
	(b) Our approach aligns natural language queries to regions that compose the candidate moment.
	}
	\label{fig:teaser}
	\vspace{-.3cm}
\end{figure*}

Consider the natural language query shown in Figure~\ref{fig:problem_statement}. 
Recent work has introduced the task of natural language moment retrieval in videos~\cite{gao2017tall,hendricks2017localizing}, where the goal is to return a relevant moment in a single untrimmed and unsegmented video corresponding to a given natural language query. 
While current methods retrieve moments from a single video, users often have large stores of untrimmed, unsegmented videos that they want to query.
\noindent For example, video content creators (\eg, professional editors) often sift through databases of raw video footage to produce a final edit. Yet, manually annotating video content is tedious and time consuming (\eg, see this tutorial~\footnote{\url{https://nofilmschool.com/premiere-pro-tagging-workflow}}). Another application is searching through surveillance or archival footage (\eg, news footage), where often the videos are long, untrimmed, and unsegmented. \\
\indent In this paper, we propose the task of temporally localizing relevant moments in a large {\em corpus} of videos given a natural language query, which could enable applications in video search and retrieval, such as video editing and surveillance.  \\
\indent Our task is challenging as we need to efficiently and accurately find both the video and exact moment in the video that aligns with a natural language query. 
While one could attempt to scale prior approaches for localizing a relevant moment in a single, untrimmed video given a natural language query~\cite{Chen18,gao2017tall,hendricks2017localizing,hendricks2018emnlp,liu2018tmn,Liu2018AttentiveMR} to a large video corpus, such an attempt would face two difficulties. 
First, we need the ability to index and efficiently retrieve relevant moments in videos. 
As current efficient indexing techniques rely on approximating the Euclidean distance between descriptors~\cite{Ge2014OptimizedPQ,Gray1998QuantizationI,Jgou2011ProductQF}, they cannot be readily plugged into video moment retrieval systems that rely on computing similarities using, often complicated, neural network architectures~\cite{Chen18,gao2017tall,liu2018tmn,Liu2018AttentiveMR}.
Second, the index size  needs to scale efficiently relative to the size of the video corpus. 
While the Moment Context Network (MCN)~\cite{hendricks2017localizing} allows for efficient retrieval due to the model's use of Euclidean distance for comparing language and video features, it requires indexing and storing all possible-length moments in a video. 
Such a requirement yields large and non-practical video index sizes. 
While indexing only action proposals~\cite{EscorciaHNG16,Gao2018CTAPCT,Xu_2017_ICCV} may be a solution to reducing the index size, such methods may discard relevant moments that a user may want to query. 

In this work, we propose \modelname{} (\modelnameabbr{}), a model that represents a video moment as a set of regions within a series of short video clips and aligns a natural language query to the moment's regions. 
Our approach is illustrated in Figure~\ref{fig:overview}. 
Our alignment cost compares variable-length language and video features using symmetric squared-Chamfer distance, which allows for efficient indexing and retrieval of the video moments. 
Moreover, aligning language features to regions within a video moment allows for finer alignment compared to methods that extract only an aggregate feature from the entire video moment.  
At query time, we propose a two-stage approach consisting of efficient retrieval followed by more expensive re-ranking to maintain recall accuracy. 
We achieve efficiency by an approximate strategy that retrieves relevant candidate \textbf{clips} for a language query using efficient approximate nearest neighbour search. 
Then, for re-ranking, we apply the full alignment cost on all variable-length \textbf{moments} in the temporal proximity of the retrieved candidate clips. 
Furthermore, representing moments as a series of short video clips allows us to overcome the need for indexing all possible variable-length moments while at the same time retrieveing any possible moment in a video. 

\paragraph{Contributions.}
Our technical contributions are twofold: we propose (i) the task of natural language video corpus moment retrieval, and (ii) a simple, yet effective, approach (\modelnameabbr{}) that aligns video~regions to a language query while allowing for efficient retrieval followed by re-ranking in the large-scale video corpus moment retrieval setting.

Our new task is interesting, practical, and challenging, and we show it to be less sensitive to the moment frequency bias issue, which hampered the progress in previous video-language moment retrieval tasks (as shown in Sec. \ref{sec:vcmr_setup}). 
Our alignment of regions in space or time to language queries is novel and conceptually different from the prior art. As discussed in Section~\ref{sec:approach}, the prior art either compares a language query feature against a fixed-length feature aggregated over a video moment~\cite{gao2017tall,hendricks2017localizing} or returns an alignment cost with a joint (neural network) model over video and language features~\cite{Chen18,ChenMCJL19,gao2017tall,liu2018tmn,Liu2018AttentiveMR,Xu19}. Based on our principled problem formulation, we derive several model instances, STAL (TEF), STAL (Clips), Approx STAL, that perform the alignment. We compute the alignment cost via the Chamfer distance, which is relatively less explored in the context of video-language understanding tasks; our work opens up the possibility for the use of Chamfer matching in other video understanding problems. Our work also showcases the relevance of the InfoNCE loss~\cite{gutmann2010_nce,VdoordLV2019_infonce} during training for video-language temporal alignment.

We are the first to perform a set of experiments in a controlled setting that explores the trade-off between retrieval accuracy and computational/memory complexity for the video corpus moment retrieval task.
We achieve this goal by extending two datasets to the video corpus retrieval setting,  DiDeMo~\cite{hendricks2017localizing} and Charades-STA~\cite{gao2017tall}, and using MLP/LSTM building blocks for representing the video and language modalities.
We demonstrate the effectiveness of our approach by showing that our \modelnameabbr{} model in an exhaustive setting out-performs MCN~\cite{hendricks2017localizing} on all recall settings across all datasets, yielding a $37$\%-$118$\% and up to $30$\% boost for average recall and median rank, respectively.  
We also out-perform a two-stage baseline of efficient video retrieval (MEE~\cite{Miech2018}) followed by re-ranking with a recent state-of-the-art approach for single video moment retrieval (2D-TAN~\cite{zhang2020learning}, CBP~\cite{WangMJ2020}, and~\cite{Xu19}).
Furthermore, for a hypothetical corpus of 1M videos, we achieve more than $130\times$ faster retrieval and $8\times$ smaller index size over MCN in an approximate setting.
Our approach obtains the best trade-off between recall accuracy and query time / memory complexity, making STAL a viable solution for the video corpus moment retrieval task at scale.

\section{Related work}
\label{sec:related_work}
Our work lies at the intersection of natural language processing and video, an area that has received much recent attention. 
Our work is closest to the tasks of, given a natural language query, retrieving short video clips from a large collection and localizing moments in a single untrimmed, unsegmented video. 
We describe related work for both tasks. 

\paragraph{Video clip retrieval with natural language.}
Recently, datasets of short video clips with accompanying natural language have emerged, including the MPII movie description dataset as part of the large scale movie description challenge (LSMDC)~\cite{Rohrbach2015ADF} and the MSR-VTT dataset~\cite{Xu2016MSRVTTAL}. 
Example recent approaches leverage detected concepts in videos~\cite{Yu2017EndtoEndCW}, hierarchical alignment and attention~\cite{Yu2018AJS}, learning a mixture of embedding~\cite{Miech2018} or collaborative~\cite{LiuANZ2019} experts, and dual deep encoding for zero-example retrieval~\cite{Dong18}. 
However, these approaches do not search for moments within untrimmed, unsegmented videos.

\paragraph{Localizing moments in a single video with natural language.} 
Datasets of videos with temporally aligned text~\cite{gao2017tall,hendricks2017localizing,hendricks2018emnlp,krishna2017dense,MiechZATLS2019,regneri2013grounding} have been used for aligning movie scripts, textual instructions, and sentences in a paragraph with a single video~\cite{Bojanowski2015WeaklySupervisedAO,Miech2017LearningFV,Shao18,Zhang18}, video object segmentation~\cite{Khoreva18}, and retrieving moments in a single video given a text query~\cite{Chen18,ChenMCJL19,chenhierarchical,gao2017tall,GaoDSX2019,GhoshAPH2019,hendricks2017localizing,hendricks2018emnlp,liu2018tmn,liu2020jointly,Liu2018AttentiveMR,MithunPR19,WangMJ2020,WangHW2019,ZhangDWWD19,zhang2020learning,ZhangSL2019,zhang2019cross}.
Our work is closest to the latter. 
As we will discuss in Section~\ref{sec:approach}, the MCN~\cite{hendricks2017localizing} and CTRL~\cite{gao2017tall} models aggregate features over a video moment before comparing to a feature for the language query.  
Our spatiotemporal alignment approach allows for finer alignment between the moment and query. 
More recent approaches have integrated alignment of clips with language queries inside a neural network as part of a temporal modular network~\cite{liu2018tmn,ZhangSL2019}, joint alignment with temporal attention~\cite{Chen18,ChenMCJL19,Liu2018AttentiveMR,Xu19}, localization via query-guided proposals~\cite{ChenJ2019,Xu19,ZhangDWWD19}, or using of weak-supervision~\cite{chen-etal-2019-weakly,GaoDSX2019,VLANet_ECCV_20,MithunPR19,song2020weaklysupervised,SCN_2020_AAAI}. 
Similar to us are approaches that align natural language to~regions in video for captioning~\cite{ZhouKCCR2019} and single video moment retrieval~\cite{WangHW2019}, with the latter using a reinforcement learning objective. As we will show, these approaches are not amenable to efficient search and retrieval at large scale. 
Our approach overcomes both limitations and allows for efficient indexing and retrieval over large video collections.
Finally, our solution is conceptually different from approaches that perform localization followed by regression~\cite{gao2017tall,GhoshAPH2019,lu_etal_2019_debug,Mun_2020_CVPR,Xu19,Zeng_2020_CVPR}. 
We efficiently retrieve candidate moments from a video corpus and re-rank them based on spatiotemporal alignment.

Similarly,~\cite{Shao18} adopts a two-stage approach for the text-to-video retrieval task where the goal is to retrieve entire videos given a natural language query.~\cite{Shao18}~coarsely ranks entire relevant videos (as in MEE~\cite{Miech2018}) based on a similarity score computed between a single video feature representation and a single language feature, followed by re-ranking. The re-ranking stage computes a sentence localization in the video and uses the grounding scores for enhancing the text-to-video retrieval scores. Differently from~\cite{Shao18}, we aim at retrieving and localizing video moments in a large corpus of videos with textual queries.

Finally, it is worth mentioning that our work has inspired new lines of research. For example, Lei \etal~\cite{lei2020tvr} extended our task to TV shows with subtitles, enabling simultaneous retrieval and localization within multimodal data~\cite{lei2020tvr,li-etal-2020-hero}.
Another example includes Otani \etal~\cite{otani2020challengesmr}, where they built on our observation regarding existing dataset bias for the single video moment retrieval task~\cite{gao2017tall,hendricks2017localizing} and investigated this bias over multiple models for the task.

\newcommand{\videocorpus}{\mathcal{V}}
\newcommand{\languagequery}{q}
\newcommand{\video}{v}
\newcommand{\temporalendpoints}{\tau}
\newcommand{\temporalstartpoint}{\tau^{(S)}}
\newcommand{\temporalendpoint}{\tau^{(E)}}
\newcommand{\clip}{c}
\newcommand{\moment}{m}
\newcommand{\featurevideo}{f^{(\video)}}
\newcommand{\featurevideoclips}{\featurevideo_{\textrm{Clips}}}
\newcommand{\featurevideoobjects}{\featurevideo_{\textrm{Obj}}}
\newcommand{\featurelanguage}{f^{(\languagequery)}}
\newcommand{\momentendpoints}{\tau^{(M)}}
\newcommand{\distancefunction}{\Phi}
\newcommand{\composingfunction}{\Psi}
\newcommand{\featurecontext}{f^{(C)}}
\newcommand{\featurecomposed}{\psi}
\newcommand{\costalignment}{\mathcal{C}}
\newcommand{\costalignmentclips}{\costalignment_{\textrm{Clips}}}
\newcommand{\costalignmentobjects}{\costalignment_{\textrm{Objects}}}
\newcommand{\videotemporalendpoints}{\mathrm{T}^{(\video)}}
\newcommand{\triplet}{\Gamma}

\begin{figure*}[t]
	\centering
	\begin{subfigure}[b]{0.45\linewidth}
	\centering
    \includegraphics[width=\linewidth]{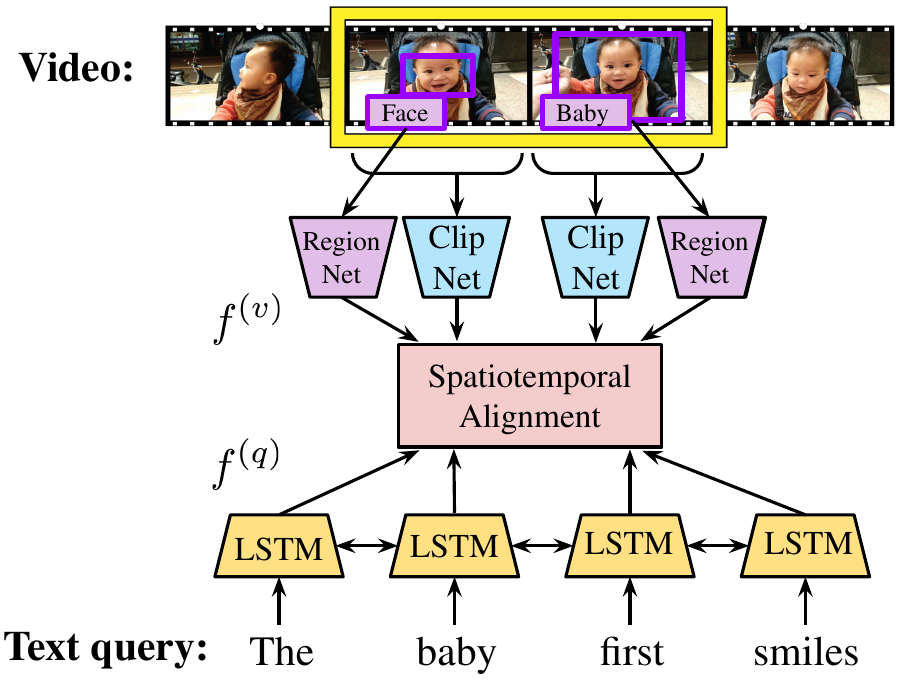}
    \caption{STAL (TEF) model architecture overview \label{fig:architecture}}
    \end{subfigure}
    \vrule
	\begin{subfigure}[b]{0.53\linewidth}
	\includegraphics[width=\linewidth]{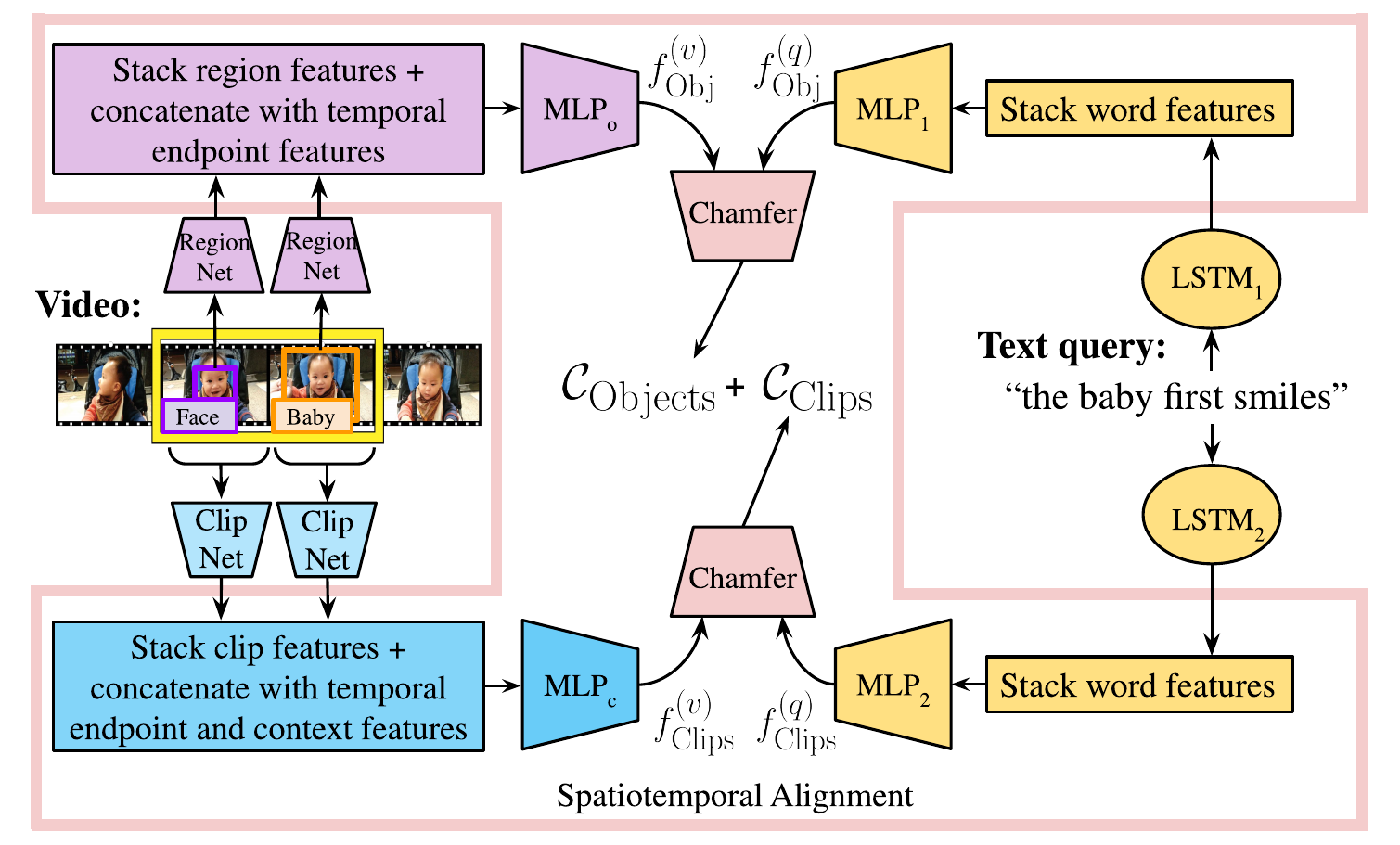}
    \caption{STAL (TEF) model with details of spatiotemporal alignment. \label{fig:model-details}}
    \end{subfigure}
	\caption{
	{\bf \footnotesize  Our SpatioTemporal Alignment with Language (STAL) model }  \footnotesize
	(a) at a glance and (b) with additional details of the spatiotemporal alignment. 
Our model is a neural network comprising two branches that align clip features $f^{(v)}_{Clips}$ and object features $f^{(v)}_{Obj}$ extracted from the video $v$ to the corresponding word representations $f^{(q)}_{Clips}$ and $f^{(q)}_{Obj}$ of language query $q$ via the evaluation of the alignment cost~(\ref{eqn:cost_alignment_chamfer}). TEF refers to temporal endpoint features that encode the position of the clip in the video.
	See text for details.
	}
	\label{fig:model}
\end{figure*}

\section{\modelname{}}
\label{sec:approach}

Our goal is, given a natural language query $\languagequery$, to return a video $\video\in\videocorpus$ from a corpus $\videocorpus$ and temporal endpoints $\temporalendpoints=\left(\temporalstartpoint,\temporalendpoint\right)$ that temporally localize the language query in the video where $\temporalstartpoint$ and $\temporalendpoint$ are start and end points, respectively. 
We model the video $\video$ as an ordered sequence of $N_c$ short
uniform-length clips $\video\in \{c_1, \cdots, c_{N_c}\}$, with corresponding temporal endpoints  $\videotemporalendpoints=\left\{\temporalendpoints_1,\cdots,\temporalendpoints_{N_c}\right\}$, and a moment $\moment^{(\video)}\subseteq\video$ as a sequence of consecutive clips $\moment^{(\video)}=\left\{\clip_{i},\cdots,\clip_{j}\right\}$ for $1\leq i \leq j \leq N_c$ with moment temporal endpoints $\temporalendpoints_\moment = \left(\temporalendpoints_{i},\temporalendpoints_{j}\right)$ for $\temporalendpoints_i, \temporalendpoints_j\in\videotemporalendpoints$. 
If the video corpus $\videocorpus$  comprises a single video, then  the task is {\em single video moment retrieval} (as proposed in \cite{gao2017tall,hendricks2017localizing}). If it is a collection of videos, the task is {\em video corpus moment retrieval} (our proposed task). 
Our approach for the video corpus moment retrieval task solution consists of two stages -- efficient retrieval followed by more expensive re-ranking by spatiotemporal alignment. We first describe our \modelname{} (\modelnameabbr{}) model and then describe how it is used for efficient two-stage retrieval with re-ranking (Section~\ref{sec:reranking}).

We cast the spatiotemporal localization problem as one of retrieving a relevant video moment $\moment^{(\video)}\subseteq\video$ from a video $\video$ with moment temporal endpoints $\temporalendpoints_\moment$ that closely correspond to the ground truth temporal endpoints $\temporalendpoints_\languagequery$ for input natural language query $\languagequery$. 
Our spatiotemporal localization problem can be formulated as an optimization over an alignment cost $\costalignment$,
\begin{equation}
    \min_{\video\in\videocorpus,\moment^{(\video)}\in\video} 
    \costalignment\mathopen{}\left(\moment^{(\video)},\languagequery\right)\mathclose{},
\end{equation}
where we aim to find the best video $\video$ and moment $\moment^{(\video)}$ that minimize the spatiotemporal alignment cost between the moment and the language query $\languagequery$. 

In this work, cost $\costalignment$ aligns two variable-length sets of features extracted over clips and regions in a video and words in a sentence, with the goal of matching objects and actions to parts of a sentence query. 
Figure~\ref{fig:overview} illustrates our overall approach and Figure~\ref{fig:model} gives the details of the model architecture. 
Let $\featurevideo = \left\{\featurevideo_1,\cdots,\featurevideo_{M_\video}\right\}$ be a set of clip and region features of size $M_\video$ extracted from a video moment $\moment^{(\video)}$ and $\featurelanguage = \left\{\featurelanguage_1,\cdots,\featurelanguage_{M_\languagequery}\right\}$ be a set of features of size $M_\languagequery$ for the language query $\languagequery$. 
As we may have a variable number of features extracted from the video moment and language query, we define the spatiotemporal alignment cost for the moment and query as the symmetric squared-Chamfer distance between the language and moment features, 
\begin{equation}
    \begin{split}
    \mathcal{C}\mathopen{}\left(\moment^{(\video)},\languagequery\right)\mathclose{} & = 
    \frac{1}{M_\video}\sum_{i=1}^{M_\video} \min_{j=1,\cdots,M_\languagequery} \left|\featurevideo_i - \featurelanguage_j\right|^2 \\
    & + \frac{1}{M_\languagequery} \sum_{j=1}^{M_\languagequery} \min_{i=1,\cdots,M_\video} \left|\featurevideo_i - \featurelanguage_j\right|^2.
    \end{split}
    \label{eqn:cost_alignment_chamfer}
\end{equation}

The Chamfer distance is the average distance of each point in one set to its closest point in the other set and has been used to align two points sets~\cite{Barrow77}. 
The first term in cost~(\ref{eqn:cost_alignment_chamfer}) finds the best matching word in the query sentence for each object candidate region or action clip and computes the average of these minimal distances. 
The second term in cost~(\ref{eqn:cost_alignment_chamfer}) computes the best matching object candidate region or action clip for each word in the text query and averages the resulting distances. 
We provide a high-level overview of our model in Figure~\ref{fig:architecture} and provide additional details of the spatiotemporal alignment in Figure~\ref{fig:model-details}.
As shown in Figure~\ref{fig:model-details}, we compute cost~(\ref{eqn:cost_alignment_chamfer}) for clip-query alignment $(\costalignment_{\textrm{Clips}})$ and object-query alignment $(\costalignment_{\textrm{Objects}})$. The final alignment cost is the sum of the two alignment costs computed over clips and objects.
Note that, while earth mover's distance~\cite{Peleg89} could be used for point set alignment, we opted for Chamfer distance due to computational complexity. 

Our alignment cost has two desirable properties for our proposed task. 
First, our cost allows finer aligment as it is an explicit function of clips and regions. 
Second, the features are indexable since the terms in the cost are Euclidean distances, which can be computed efficiently using approximate nearest neighbor techniques such as~\cite{Jgou2011ProductQF,MujaL09}. We discuss next the advantages of both properties in relation to prior work.

\paragraph{Discussion.} 
Prior work~\cite{gao2017tall,hendricks2017localizing} compares a single fixed-length language feature $\featurelanguage$ to a fixed-length feature aggregated over the video moment,
\begin{equation}
    \mathcal{C}_{\textrm{agg}}\mathopen{}\left(\moment^{(\video)},\languagequery\right)\mathclose{} =     
    \distancefunction\mathopen{}\left(\composingfunction\mathopen{}\left(\featurevideo_1,\cdots,\featurevideo_{M_\video}\right)\mathclose{},\featurelanguage\right)\mathclose{},
\end{equation}
where $\composingfunction$ aggregates $M_\video$ clip features $\featurevideo_1,\cdots,\featurevideo_{M_\video}$ into an embedded feature for the candidate moment and $\distancefunction$ compares the aggregated video moment and language features. 
In MCN~\cite{hendricks2017localizing}, squared-Euclidean distance $\left(\distancefunction\right)$ is used to compare aggregated video moment $\left(\composingfunction\right)$ and language features. 
In CTRL~\cite{gao2017tall}, aggregated video moment features $\left(\composingfunction\right)$ and language features pass through a neural network $\left(\distancefunction\right)$. 
One drawback of these formulations is that the language feature is compared to an aggregated feature over the entire moment and does not have the ability to align to the individual regions in the moment. 

In recent work~\cite{Chen18,ChenMCJL19,gao2017tall,liu2018tmn,liu2020jointly,Liu2018AttentiveMR,Xu19,zhang2020learning,zhang2019cross}, a joint model over language and video features is used to return the alignment cost,
\begin{equation}
    \mathcal{C}_{\textrm{joint}}\mathopen{}\left(\moment^{(\video)},\languagequery\right)\mathclose{} =     
    \distancefunction\mathopen{}\left(\left(\featurevideo_1,\cdots,\featurevideo_{M_\video}\right),\featurelanguage\right)\mathclose{},
\end{equation}
where $\distancefunction$ is a neural network, possibly encompassing a learnable distance function, that outputs a scalar value of the cost of aligning the video features set $\left\{\featurevideo_1,\cdots,\featurevideo_{M_\video}\right\}$ with a language query feature $\featurelanguage$.
While these approaches have achieved early success for single video moment retrieval, they currently cannot perform efficient indexing and retrieval at a large scale (\eg, over millions of untrimmed and unsegmented videos) due to their reliance on a neural network for comparing video and language features,~\ie, it would be too expensive to compute the alignment cost at test time for a large video corpus.

\paragraph{STAL (TEF) model details.} Here we give the details of our spatiotemporal alignment with language (STAL) model.
Figure~\ref{fig:model-details} illustrates our full model. 
A set of video features $\featurevideo$ is computed and compared to a set of language features $\featurelanguage$ for query $\languagequery$ using Chamfer distance. 
We compute two types of video features -- one set over the short video clips that aims to capture the overall content of the video including human actions
$\featurevideoclips$, and the other over detected objects that aims specifically to capture the properties of objects present in the video
$\featurevideoobjects$. For each clip feature, we concatenate  visual  features computed over the temporal extent of the clip with a context feature and temporal endpoints for the moment, which are then passed through a multilayer perceptron (MLP). 
As in MCN~\cite{hendricks2017localizing}, for context features, we average pool clip features over the entire video.
The intuition is to enrich the local information of each clip with global information about the content of the video.
For the detected object features, we concatenate the word embedding of the detected object name with the spatiotemporal location of the detected object, which are then passed through an MLP. We investigate benefits of alternative object representations in appenddix~\ref{sec:object-region-features}.
The language features $\featurelanguage$ are the outputs of the hidden layer of an LSTM for each query word's embedded feature, followed by a linear mapping. 
We use pre-computed features for the visual and word-embedding features and pre-trained object detectors, so our model parameters comprise the two MLPs, LSTM, and hidden layer linear mapping. 
As we will show in Section~\ref{sec:experiments}, computing cost~(\ref{eqn:cost_alignment_chamfer}) separately over object ($\costalignmentobjects$) and clip features ($\costalignmentclips$) and jointly training both branches performs best.

For a fair comparison with MCN, we have built on their architectural design and have chosen a simple model for the visual stream, which does not explicitly model relations between the individual image regions or the clips in the video. However, our model aligns the regions and the clips of the video to language features of the query using a bidirectional LSTM. As a result, each word is mapped and aligned in the context of the whole sentence. The LSTM model represents the ordering of the words in the query sentence and can distinguish queries like ``\textit{the man chases the dog}'' and ``\textit{the dog chases the man}'', which would result in different language representations. Incorporating more advanced visual features that take into account the relationships across the visual regions is an exciting direction of future work.

For the context feature, as discussed above, we adopt the same modeling strategy as MCN~\cite{hendricks2017localizing} so that we can directly compare our methods with theirs. Note that this modelling strategy does not explicitly represent information from neighbouring clips, which is done, \eg, in CTRL~\cite{gao2017tall}. In contrast, it captures the context by aggregating features over the entire video.

\newcommand{\trainingloss}{\mathcal{L}}
\newcommand{\nceloss}{\trainingloss^{\textrm{NCE}}}
\newcommand{\tripletloss}{\trainingloss^{\textrm{Triplet}}}
\newcommand{\rankingloss}{\mathcal{L}^R}
\newcommand{\modelparameters}{\theta}
\newcommand{\trainingset}{\mathcal{P}}
\newcommand{\negativesetintra}{\mathcal{N}_\textrm{intra}}
\newcommand{\negativesetinter}{\mathcal{N}_\textrm{inter}}
\newcommand{\allnegatives}{\mathcal{N}}
\newcommand{\interparameter}{\lambda}
\newcommand{\costalignmenttrain}{\Tilde{\mathcal{C}}}

\paragraph{Training.} 
We seek to have our \modelnameabbr{} model rank moment-query pairs such that aligned pairs are pushed to the top of the list while misaligned pairs are pushed to the bottom. To achieve this goal, we define a Noise Contrastive Estimation (NCE) training objective~\cite{gutmann2010_nce,VdoordLV2019_infonce}.

Let $\trainingset=\left\{\left(\moment^{(\video)},\languagequery\right)_k\right\}_{k=1}^{N_v}$ be a training set of $N_v$ aligned video moment and natural language query pairs. 
For a positive training example $p\in\trainingset$, we define an intra-video negative set $\negativesetintra^{(p)}$ consisting of video moments in the training example video not aligned to the language query training example. 
Similarly, we define an inter-video negative set $\negativesetinter^{(p)}$ consisting of video moments from completely different videos in the training set.
For a positive training example sampled from the training set $p\sim\trainingset$, let $n$ be samples from the intra- and inter-video negative sets, $n \in \allnegatives^{(p)}$ with $\allnegatives^{(p)} = \negativesetintra^{(p)} \cup \negativesetinter^{(p)}$.
Let $\costalignmenttrain_p = \costalignment\mathopen{}\left(\moment^{(\video)},\languagequery\right)\mathclose{}$ be the alignment cost~(\ref{eqn:cost_alignment_chamfer}) for positive training example $p=\left(\moment^{(\video)},\languagequery\right)$.
Similarly, let $\costalignmenttrain_n=\costalignment{\left(n,\languagequery\right)}$ represent the negative alignment cost, computed for negative moment $n$.
We formulate the InfoNCE loss as the ratio of the positive-sample distance to the sum of positive and negative (intra and inter) sample distances,
\begin{equation}
    \nceloss_\modelparameters =  - \sum_{p\in\trainingset} \log{\left(
    \frac{
     \exp{\left(-\costalignmenttrain_p\right)}
     }
    {\exp{\left(-\costalignmenttrain_p\right)}+
     \underset{n\in \allnegatives^{(p)}}{\sum} \exp{\left(-\costalignmenttrain_n\right)}
     } \right)}.
    \label{equation:contrastive}
\end{equation}

This loss encourages the Chamfer distance of the positive sample $\costalignmenttrain_p$ to be smaller than the Chamfer distance to all the negative samples.
We optimize the training loss $\nceloss_\modelparameters$ for model parameters $\modelparameters$, using stochastic gradient descent with momentum by uniform sampling over positive, intra-, and inter-negative sets.

\begin{figure*}[t]
	\centering
	\begin{subfigure}[b]{0.49\linewidth}
	\centering
    \includegraphics[width=\linewidth]{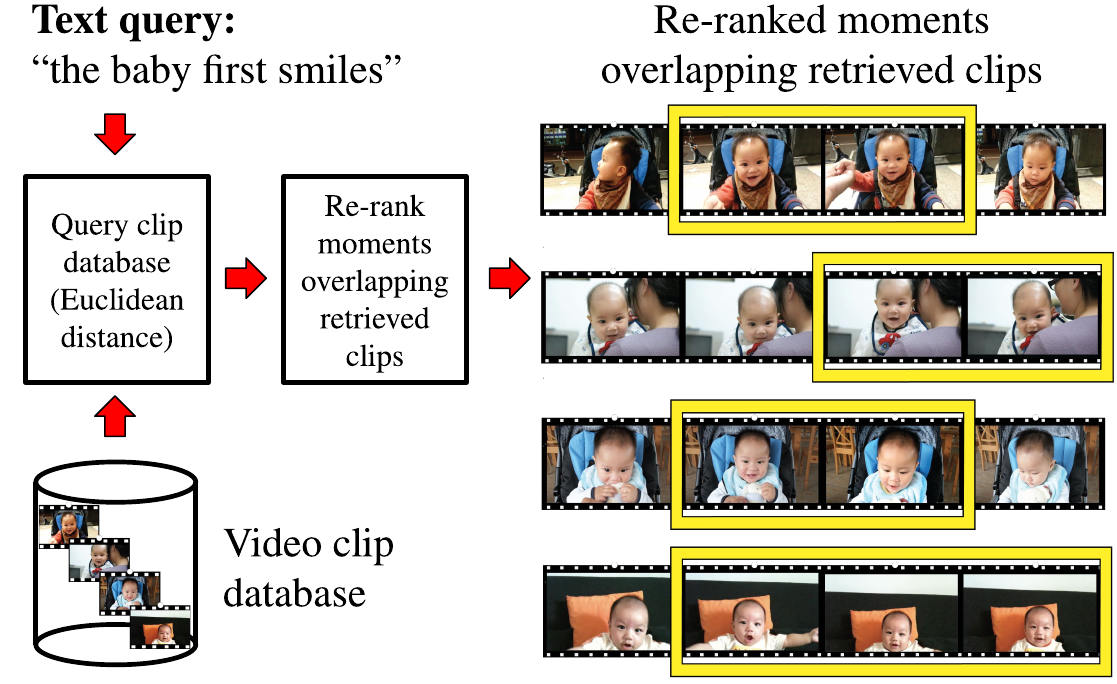}
    \caption{Efficient retrieval with re-ranking \label{fig:retrieval_overview}}
    \end{subfigure}
    \vrule
	\begin{subfigure}[b]{0.49\linewidth}
    \includegraphics[width=\linewidth]{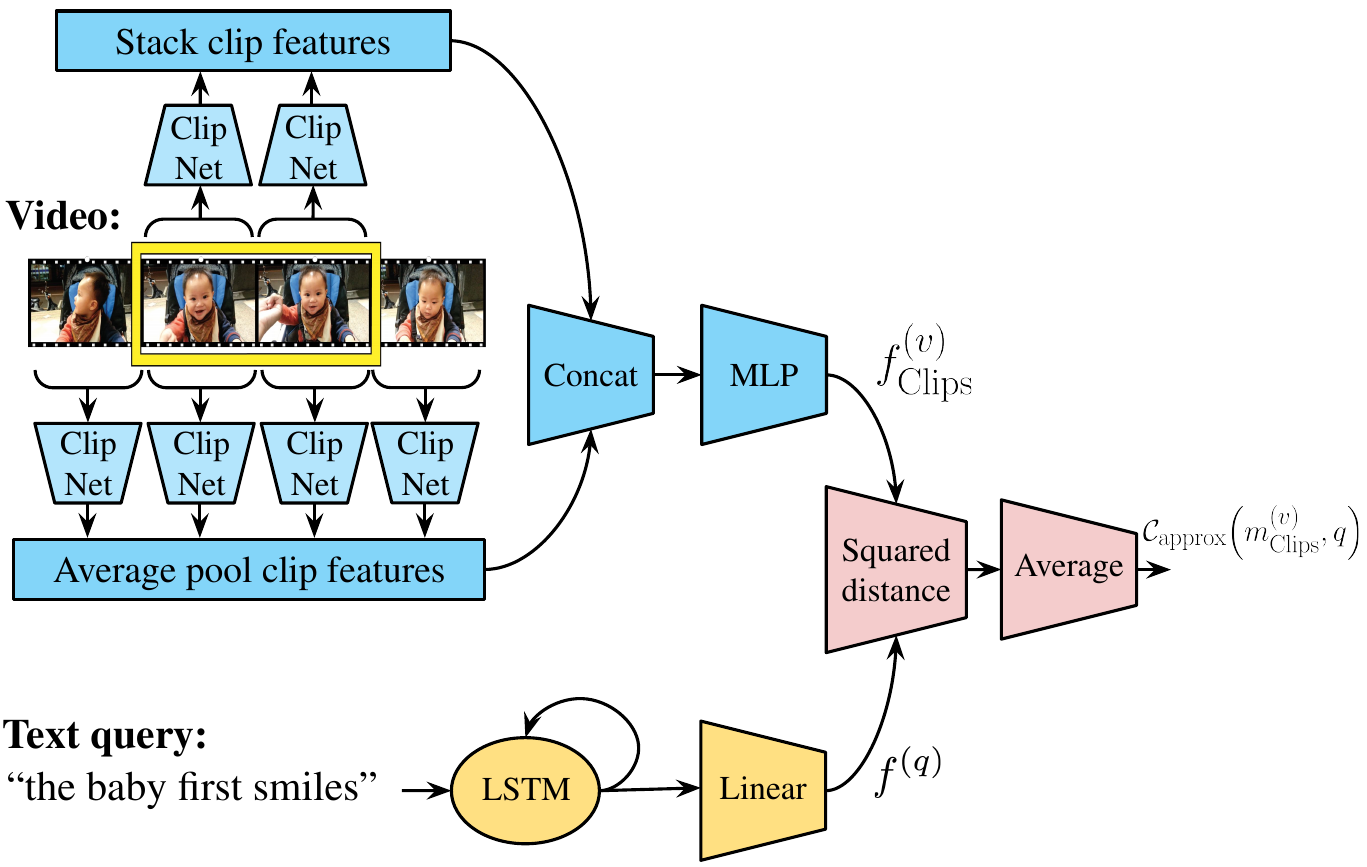}
    \caption{ STAL (clips) architecture.  \label{fig:stal-clips}}
    \end{subfigure}
	\caption{
	\footnotesize
	{\bf System for indexing and retrieval.}
	(a) Our approach allows for efficient storage and retrieval of variable-length moments in a video collection database via a two-stage approach.
	The first stage, efficient retrieval, uses a lightweight version of our model, termed (b) ``STAL (clips)''. Clip features $f^{(v)}_{\text{ Clips}}$ are matched to the query vector $\featurelanguage$ by means of Euclidean distance.
	We train ``STAL (clips)'' embedding with the cost~(\ref{eqn:cost_alignment_euclidean}).
	The second stage, re-ranking, ranks all possible moments of different lengths containing the top retrieved clips from the first stage, using the cost~(\ref{eqn:cost_alignment_chamfer}).
	See text for details.
	}
	\label{fig:models_details}
\end{figure*}

\subsection{Efficient retrieval with re-ranking}
\label{sec:reranking}

For inference, one can evaluate cost~(\ref{eqn:cost_alignment_chamfer}) exhaustively over all possible moments in all videos. 
While this routine was used in MCN~\cite{hendricks2017localizing} to localize moments in a single video, this exhaustive strategy does not efficiently scale to localizing moments in a large video corpus. 
To achieve efficient retrieval while maintaining recall accuracy, we propose a two-stage approach consisting of an efficient retrieval stage followed by a more expensive re-ranking stage. 

Our \modelnameabbr{} model allows for efficient indexing and retrieval of video moments for a natural language query since it relies on comparing video and language features with Euclidean distances. 
This is important for our application as we may potentially want to search through a large corpus comprising millions of untrimmed, unsegmented videos. 
As noted in our earlier discussion, approaches that align video and language features with neural networks currently do not extend to large-scale indexing applications, which is a key difference from our approach. 

Our strategy for implementing the efficient retrieval stage with our approach is to index a set of all video clip features for each video. 
At query time, the system retrieves moments in a greedy fashion by retrieving the top clips corresponding to a single language feature for a query.
All possible moments of different lengths that include the top retrieved clips are then ranked using the more expensive alignment cost. In detail, 
for efficient retrieval we introduce ``\modelnameabbr{} (clips)'', a lightweight version of our STAL (TEF) model that is used in the efficient retrieval stage (see Figure~\ref{fig:stal-clips}).
Let $\featurevideo_{\textrm{Clips}} = \{\featurevideo_1,\cdots,\featurevideo_Z\}$ be the set of clip features within the moment and $\featurelanguage$ be a single feature vector for the language query $\languagequery$. We approximate our alignment cost as the average squared-Euclidean distance between the language feature and the moment's clip features, 
\begin{equation}
    \mathcal{C}_{\textrm{approx}}\mathopen{}\left(\moment_{\textrm{Clips}}^{(\video)},\languagequery\right)\mathclose{} =     
    \frac{1}{Z}\sum_{k=1}^{Z} \left\|\featurevideo_k-\featurelanguage\right\|^2,
    \label{eqn:cost_alignment_euclidean}
\end{equation}
where $Z>1$ is the number of clips in the moment. Figure \ref{fig:stal-clips} shows the block diagram of our lightweight \modelnameabbr{} (clips) model. In our implementation, $\featurelanguage$ corresponds to the last hidden state of the LSTM for the query. 

We train the ``\modelnameabbr{} (clips)'' model with cost~(\ref{eqn:cost_alignment_euclidean}), which is an approximation of cost~(\ref{eqn:cost_alignment_chamfer}).
For the re-ranking stage, we score and re-rank the set of moments containing the top retrieved clips with the more expensive cost~(\ref{eqn:cost_alignment_chamfer}) via the \modelnameabbr{} (TEF) model. 
This strategy is illustrated in Figure~\ref{fig:retrieval_overview}. 
While this efficient retrieval with re-ranking strategy is not guaranteed to retrieve the best moment in terms of cost~(\ref{eqn:cost_alignment_chamfer}), we are able to effectively return the correct moment in practice (see Section~\ref{sec:experiments}). 
Moreover, our approach allows for retrieval of any moment from any video, which is in contrast to proposal-based methods~\cite{Gao2018CTAPCT} that discard clips from videos. 

\begin{table*}[!t]
\caption{
\footnotesize 
{\bf STAL architecture variants.} 
We consider different architectures for efficient retrieval followed by re-ranking. 
 ``a. STAL (clips)'' is a lightweight variant of our model used for fast retrieval of candidate clips. It uses only clip features and a simplified alignment cost~\eqref{eqn:cost_alignment_euclidean} based on the Euclidean distance. ``b. Approx STAL'' is a faster version for large datasets using an efficient approximate nearest neighbour search.  ``c. STAL (TEF)'' is our full model using clip, object, and temporal endpoint (TEF) features, and is used for re-ranking candidate clips returned by models a. or b. STAL (TEF) retrieves video moments of variable length by performing spatiotemporal alignment using cost~\eqref{eqn:cost_alignment_chamfer}.
For index size, let $M$ be the number of all possible moments in a corpus, $N$ be the number of clips in the corpus, and $d$ be the embedding dimensionality. Note that $M$ can be much larger than $N$. Please see the text for more details.
} 
\centering 
\footnotesize
\scalebox{0.97}{

\begin{tabular}{l|c|c|c|c|c|c|c|c}
\toprule

& \multicolumn{3}{c|}{ Features    }
& Cost      
& Index Size 
& Retrieves   
& Search
& Pipeline 
\\

& \multirow{1}{*}{Clip}
& \multirow{1}{*}{Obj.}
& \multirow{1}{*}{TEF }
&  
& 
& 
& Method
& Stage
\\

 \midrule

  a. STAL (clips)    
& \cmark      
& \xmark      
& \xmark      
& Cost eq.~(\ref{eqn:cost_alignment_euclidean})  
& $\mathcal{O}(Nd)$ 
& Clips             
& Exhaustive       
& Efficient retrieval       
\\

b. Approx. STAL  
& \cmark     
& \xmark     
& \xmark     
& Cost eq.~(\ref{eqn:cost_alignment_euclidean}) 
& $\mathcal{O}(Nd)$ 
& Clips             
& Approximate 
& Efficient retrieval       
\\

  c. STAL (TEF)  
& \cmark      
& \cmark      
& \cmark      
& Cost eq.~(\ref{eqn:cost_alignment_chamfer}) 
& $\mathcal{O}(Md)$ 
& Moments          
& Exhaustive       
& Re-ranking      \\
\bottomrule
\end{tabular}
}
\label{tab:stal_arch}
\end{table*}

While MCN~\cite{hendricks2017localizing} can also index features corresponding to video moments, our approach offers an advantage with respect to the index size. For our clip-alignment approach, only $N_c$ clips are indexed for a video. 
For MCN, all possible-length moments must be indexed as the model relies on aggregated features over the moments. 
Assuming maximum moment length of $L$ clips, this results in an index of size $N_cL-\frac{1}{2}L(L-1)$ for a video.

For example, suppose we have a one-minute long video with clip length of $3$ seconds. Therefore, we have $N_c = 60\textrm{s} / 3\textrm{s} = 20$ clips in the video. Suppose we allow for a max moment length of $L=5$ clips. Plugging these quantities into the equation yields $N_cL-\frac{1}{2}L(L-1) = 20 \times 5 - 0.5 \times 5 \times (5 - 1) = 90$ possible moments in the video. Let $d$ denote the dimension of the shared visual-text embedding. As MCN requires indexing all possible moments, for this example, its index size corresponds to $90d$. Conversely, STAL indexes clips effectively reducing the index size to $20d$.
For the datasets considered in this paper, this results in $6\times$--$12\times$ decrease of the index size of our approach compared to MCN.
This difference is expected to get even larger when longer, more complex moments need to be considered, thus increasing the value of $L$.

We summarize the different STAL architectures in Table~\ref{tab:stal_arch}. STAL (clips) differs from STAL (TEF) as the former adopts the Euclidean distance (Eq.~(\ref{eqn:cost_alignment_euclidean})) instead of the more computationally demanding Chamfer distance (Eq.~(\ref{eqn:cost_alignment_chamfer})) and does not take advantage of the temporal endpoint features (TEF). Note that TEF refers to the entire moment and does not decompose into the individual clips (\eg  ~given a video containing $5$ clips, and considering the moment that span the first $2$ clips, the TEF information is expressed as a tuple of numbers: $(0.0,0.4)$, which refer to the normalized start and end of the moment with respect to the video duration).  
Moreover, STAL (clips) relies on the clip features only and not on the object features, which allows for a faster query time. As STAL (clips) exhaustively searches for relevant clips, we also consider using approximate nearest neighbour search for the first-stage retrieval (Approx.\ STAL). Both STAL (clips) and (Approx STAL) use the same trained parameters and features; the only difference is that in Approx STAL the retrieval is implemented using approximate nearest neighbour search, which results in a further increase in speed at the cost of a (small) loss in accuracy. In practice, we train separately the efficient retrieval stage, STAL (clips) / Approx STAL, from the re-ranking model, STAL (TEF). We tried out a more complex training scheme involving re-training the re-ranking stage based on the outputs of the retrieval model. However, the more complex training scheme achieves a similar retrieval accuracy. Thus, separately training our retrieval and re-ranking models is simple, yet effective and fulfills the accuracy and efficiency requirements of our task. We use the publicly available Faiss implementation~\cite{Johnson2017BillionscaleSS} to index and efficiently retrieve the clip features. Practically speaking, we adopt the IVFADC method (coarse quantizer + PQ encoding on residuals) with the default parameters. Note that our methodology is independent of the specific approximate nearest neighbour search algorithm.

\vspace{-0.45cm}
\section{Experiments} \label{sec:experiments}

In this section, we show qualitative and quantitative results on our proposed task of retrieving relevant moments from a large corpus of videos for a natural language query.
We start by describing the details of our novel task (Section~\ref{sec:vcmr_setup}).
Next, we show results on our proposed task in an exhaustive setting (Section~\ref{sec:video_corpus}).
Finally, we demonstrate results using efficient retrieval with re-ranking (Section~\ref{sec:video_corpus_twostage}).

\subsection{Video corpus moment retrieval setup}
\label{sec:vcmr_setup}

\begin{table*}[ht]
\caption{\label{tab:dataset_settings} 
{\bf \footnotesize Dataset settings and statistics.} \footnotesize Relevant statistics are computed for videos (cardinality, average duration, etc.), and natural language annotations (cardinality, measure of textual lexical diversity (MTLD)~\cite{mccarthy2005assessment}) for DiDeMo~\cite{hendricks2017localizing} and Charades-STA~\cite{gao2017tall} datasets. R-C3D proposals indicates the use of Action proposals from~\cite{Xu_2017_ICCV}. On the right, we report the oracle recall upper bound for the single video moment retrieval task. 
}
\centering 
\footnotesize
\scalebox{0.88}{
\begin{tabular}{lcccccccccc} 
\toprule
Dataset & 
Num & 
Num & 
MTLD &
R-C3D &
Clip &
Max moment &
Stride &
Avg.\ video &
\multicolumn{2}{c}{Recall} \\ 

& videos 
& 
moments & 
& 
proposals& 
length 
& 
length &
& 
length &
IoU=0.5 &
IoU=0.7 \\
\midrule
DiDeMo~\cite{hendricks2017localizing} & 10k+ & 41k+ & 66.0 & \xmark & 2.5 secs. & 6 clips & 5 secs. & 29 secs. & 100.00 & 100.00 \\
\midrule
\multirow{2}{*}{Charades-STA~\cite{gao2017tall}} & \multirow{2}{*}{6k+} & \multirow{2}{*}{16k+} & \multirow{2}{*}{17.8} & \xmark & 3 secs. & 8 clips & 3-6 secs. & 31 secs. & 99.62 & 88.79 \\
& & & & \cmark & 3 secs.  & 106.7 secs. & - & 31 secs. & 96.08 & 77.07 \\
\bottomrule

\end{tabular}
}
\end{table*}

\paragraph{Datasets.}
We evaluate on two datasets that have natural language sentences aligned in time to videos and have been proposed for the single video moment retrieval task: DiDeMo~\cite{hendricks2017localizing}, and Charades-STA~\cite{gao2017tall}. These datasets have a large number of temporally aligned natural language sentences with large (open) vocabulary. 
Moreover, the videos depict general scenes and are not constrained to a specific scene type. 
DiDeMo consists of unedited video footage from Flickr with sentences aligned to unique moments in the video (\ie, the sentences are {\em referring}). 
There are 10642 videos and 41206 sentences in the dataset and we use the published splits over videos (train\textendash8511, val\textendash1094, test\textendash1037). 
Note that moment start and end points are aligned to five-second intervals and that the maximum annotated video length is 30 seconds. 
Charades-STA builds on the Charades dataset~\cite{sigurdsson2016hollywood} consisting of unedited videos of humans acting from scripts. 
There are 6670 videos and 16124 sentences in the dataset and we use the published splits over videos (train\textendash5336, test\textendash1334). 
The videos are typically longer in length than the ones in DiDeMo and sentences from the scripts are aligned in time and may not be referring. 
Table~\ref{tab:dataset_settings} summarizes the statistics of both datasets, including the Measure of Textual Lexical Diversity (MTLD)~\cite{mccarthyJ2010_mtld} which measures the ``lexical richness'' of the textual descriptions. The higher values indicates more diversity. We observe than DiDeMo descriptions are more diverse than Charades-STA.

We adapt the DiDeMo and Charades-STA datasets used for single video moment retrieval to our video corpus moment retrieval task. 
Specifically, a method must correctly identify both the video and the moment within the video corresponding to a ground truth natural language query.

\paragraph{Evaluation criteria.} 
We adopt the criteria proposed in TALL~\cite{gao2017tall}, where average recall at $K$ (R@$K$) is reported over all language queries. 
We measure recall for a particular language query by determining whether one of the top $K$-scoring retrieved moments sufficiently overlaps with the ground truth annotation (recall will be 0 or 1). A retrieved moment sufficiently overlaps with a ground truth annotation if the ratio of the temporal intersection over union (IoU) exceeds a specified threshold. 
We average the recall values across all language queries to obtain the average recall at $K$. 
We report R@$K$ over all retrieved moments from the video corpus for $K\in\{1,10,100\}$, averaged over $\textrm{IoU}\in\{0.5,0.7\}$. 
In addition, we report the median rank for the correct retrieval, averaged over the two IoU settings. 
While the annotations are not exhaustive (\ie, a given natural language query may appear in a video but not be annotated), reporting over different values of $K$ allows us to take into account the missing annotations.
Finally, note that DiDeMo~\cite{hendricks2017localizing} has multiple annotations for each sentence corresponding to different human judgements. 
We account for the multiple annotations by requiring that a correct detection must overlap with at least two of the human judgements with the specified IoU, which can be satisfied for all sentences in the val and test sets.

\paragraph{Implementation details.} 
To obtain candidate moments in a video, we need to specify the clip length, maximum number of clips in a moment, and how frequently to extract clips in a video (temporal stride). 
Table~\ref{tab:dataset_settings} shows the settings for the video clip length, maximum moment length, and temporal stride used for the evaluated datasets.
We set the values for each dataset to maximize an oracle detector where a sequence of (non-overlapping) clips are aligned with the ground truth moments, while minimizing computational cost. 
We set the temporal stride to 5 seconds for all moments in DiDeMo and proportionally to the moment length $l_m$ in the other datasets computed as $0.3 \times l_m$ (rounded to the nearest clip boundary) as longer-length moments do not need fine temporal stride. 
Given the settings in Table~\ref{tab:dataset_settings}, the number of candidate moments for each dataset are: DiDeMo \textendash~ 21,777, Charades-STA \textendash~ 49,465.  
For approaches, such as MCN, that index all possible moments, the index size for the evaluated datasets is proportional to the number of moments.
In contrast, the index size of our efficient retrieval approach is proportional to the total number of clips in the video collection $\videocorpus$, \cf~Section~\ref{sec:reranking}. The average number of clips $N_c$ for each dataset is the ratio of the average video duration to the clip duration, which can be computed from Table~\ref{tab:dataset_settings}. The average video duration for the two datasets is approximately 30s. Given the clip duration of 5s and 3s for DiDeMo and Charades-STA, respectively, we can determine that $N_c=6$ and $N_c=10$, respectively.
We report the performance of the oracle detector in Table~\ref{tab:dataset_settings}. 
While the oracle's returned endpoints align to clip boundaries and do not have the ability to exactly align to ground truth endpoints, we note that the oracle detector still achieves high performance. 
Also note that humans may not generally agree on temporal endpoints~\cite{Alwassel18}. 
For all approaches, we evaluate their alignment cost for every moment in a video and perform non-minimum suppression with temporal IoU threshold chosen empirically for each dataset (DiDeMo \textendash~ 1.0, Charades-STA \textendash~ 0.6). We also report the oracle performance using R-C3D action proposals~\cite{Xu_2017_ICCV}. Notice that the action proposals have lower recall than the sliding window approach, making them less suited for our task. To make the comparison fair, we use the same number of proposals across the two methods, which keeps the computational cost fixed and allows for direct comparison. 

For clip features, our model uses ResNet-152 $pool5$ features~\cite{He2016DeepRL} computed over the video frames extracted at 5 fps in a clip. Then, we average pooled over the spatial dimensions, followed by max pooling over the temporal dimension, resulting in a 2048-dimensional feature for the clip. Empirically, we observed that max pooling over the temporal dimension outperformed average pooling. 
We normalized the start and end points relative to the video length as in MCN~\cite{hendricks2017localizing} to obtain the temporal endpoint features (TEFs).
For the object detection features, we ran the publicly available pre-trained object detector~\cite{Anderson2017up-down} (Faster R-CNN~\cite{renNIPS15fasterrcnn}, ResNet-101~\cite{He2016DeepRL} features) trained over the visual genome~\cite{krishnavisualgenome} at one frame per second. 
The encoding of each detected object is 306-dimensional, as we construct these features by concatenating the Glove encoding of the detected object's class label ($300$-D) with its spatial location ($4$-D consisting of the detected bounding box centroid, width, and height normalized to the image's width/height), and the TEF feature ($2$-D) of the moment of interest. 
For each clip, we consider the $10$ most salient objects given by the object detector confidence scores. 
Note that for both the clip and object features, we perform the stack operation, which forms a $N_o \times D$ matrix for $N_o$ vectors each of length $D$.

We used Glove word-embedding features~\cite{Pennington2014GloveGV} for the words in the language query and the object detection vocabulary. 
For stochastic gradient descent, we set momentum to 0.95 and used a schedule of lowering an initial learning rate of 0.05 by a factor of 0.1 every 30 epochs; training stopped at 108 epochs.
We formed mini-batches with 128 positive/negative examples.
We selected intra-negatives such that their overlap with the ground-truth moment is lower than a given IoU value.  
For DiDeMo, we used IoU=1 since the ground truth is aligned to five-second intervals; for Charades-STA we used IoU=0.35. 
Similarly, inter-negatives were selected from the same temporal location as the ground-truth moment, whenever possible, in another video selected at random from the entire dataset.

\begin{table}[!t]
\caption{\label{tab:vcmr_vs_svmr} 
{\bf \footnotesize Relevance of our proposed video corpus moment retrieval (VCMR) vs.\ the single video moment retrieval (SVMR) task~\cite{gao2017tall,hendricks2017localizing}}. \footnotesize We report the average recall for $K=1$, averaged over $\textrm{IoU}=\{0.5,0.7\}$, as well as the relative difference \wrt~the TEF-only baseline for the corresponding task and dataset in parentheses. We evaluate the models on the test set for both datasets. Refer to text for more details.}
\centering 
\footnotesize
\resizebox{0.85\linewidth}{!}{
\begin{tabular}{l|cc|cc} 
\toprule
Method & \multicolumn{2}{c|}{DiDeMo} & \multicolumn{2}{c}{Charades-STA} \\
 & SVMR~~~~~~ & VCMR~~~~~~ & SVMR~~~~~~ & VCMR~~~~~~\\ 
\midrule
Chance    & 0.12~~~~~~~  & 0.00~~~~~~~~~ & 0.12~~~~~~  & 0.01~~~~~~~ \\
TEF-only  & 24.36~~~~~~~~ & 0.04~~~~~~~~~ & 20.93~~~~~~~~ & 0.03~~~~~~~ \\
MCN       & 21.29 {\tiny(-0.13)} & 0.30 {\tiny(+6.50)} & 15.18 {\tiny(-27.4)} & 0.13 {\tiny(+3.3)} \\
MCN (TEF) & 30.50 {\tiny(+0.25)} & 1.03 {\tiny(+24.8)} & 24.65 {\tiny(+0.18)} & 0.23 {\tiny(+6.7)} \\
\bottomrule
\end{tabular}
}
\vspace{-12pt}
\end{table}

\paragraph{Relevance of the video corpus moment retrieval task}.
To further motivate the relevance of our proposed task, we assess if a model could solve our task disregarding the visual information of the videos.
Previous work in image-language understading, namely visual question answering (VQA)~\cite{Goyal2018MakingTV}, reports the existence of dataset biases allowing a model to score well enough on a given task without considering the visual information.
\cite{hendricks2017localizing} noted that TEF is an important cue for achieving good enough grounding results in DiDeMo for the single video moment retrieval (SVMR) task. Thus, we probe both datasets (\ie, DiDeMo~\cite{hendricks2017localizing} and Charades-STA~\cite{gao2017tall}), and both tasks (\ie, VCMR and SVMR) with four different baselines making use (or not) of TEF. In detail, we
train two MCN models, one with and another without TEF features. As a control method, we train a TEF-only baseline over TEF and textual descriptions only (no visual information is used).
As the TEF-only model resembles the same architecture of MCN, this baseline is the fairest comparison as it does not inject an inductive bias.
Table~\ref{tab:vcmr_vs_svmr} reports the average recall for $K=1$, averaged over $\textrm{IoU}=\{0.5,0.7\}$, for each baseline, and the relative change \wrt~the TEF-only baseline on the corresponding dataset and task in parentheses.

\noindent{\it TEF-only, a better baseline for SVMR.}~We observe that TEF-only is a strong baseline compared to Chance for the SVMR task rendering an absolute difference of $20.8-24.2\%$ percentage points. 
In contrast, the difference between TEF-only compared to Chance in our proposed VCMR task is relatively small, $0.03-0.04\%$ absolute difference, due to a larger search space.

\noindent{\it What is the effect of TEF on the datasets?}~Along the same lines of~\cite{hendricks2017localizing}, we observed that TEF-only achieves better localization results than MCN without TEF in Charades-STA. The fact that the TEF-only baseline performs so well indicates that there is a strong bias in the existing SVMR benchmarks as only knowing the language query and the relative position of the moment in the video can allow for reasonably high accuracy. This fact was also observed in early datasets for visual question answering (VQA)~\cite{Goyal2018MakingTV}.

\noindent{\it Effect of TEF on video moment corpus retrieval.}~We observed that TEF is a relevant cue to improve the performance on our proposed task (\ie, MCN (TEF) vs.\ MCN). However, TEF-only is not stronger than MCN without TEF. Thus, video-language models truly need to pay attention to the visual content of the video to solve our task.
In conclusion our proposed task not only poses a more challenging video-language problem, but also exhibits less dataset bias than the existing single video moment retrieval task~\cite{gao2017tall,hendricks2017localizing}.

\begin{table*}[!t]
\caption{
\label{tab:chamfer_ablation}
	{\bf \footnotesize Ablation study on DiDeMo (val) exhaustive setting.}
	\footnotesize{Refer to text for more details.}
}
    \centering
    \begin{subtable}[t]{.3\linewidth}\centering
    \caption{\footnotesize\textbf{Clip vs.\ moment alignment}} 
    \label{tab:ablation_alignment}
    \scalebox{0.7}{
    \begin{tabular}{lcccc}
        \toprule
        Method &  ~$K$=1 & ~$K$=10 & ~$K$=100 & ~MR $\downarrow$ \\
        \midrule
        MCN & 1.22 & 6.22 & 26.81 & 382 \\
        Clips-only & {\bf 1.38} & {\bf 6.47} & {\bf 28.05} & {\bf 357} \\
        \bottomrule
    \end{tabular}
    }
    \end{subtable}
    \begin{subtable}[t]{.3\linewidth}\centering
    \caption{\footnotesize\textbf{Spatiotemporal alignment}} 
    \label{tab:ablation_spatio_temporal}
    \scalebox{0.7}{
    \begin{tabular}{lcccc}
        \toprule
         & ~$K$=1 & ~$K$=10 & ~$K$=100 & ~MR $\downarrow$ \\
        \midrule
        Clips & 1.38 & 6.47 & 28.05 & 357 \\
        Objects & 0.53 & 3.50 & 19.41 & 629 \\
        Clips+Objects & {\bf 1.41} & {\bf 7.40} & {\bf 30.23} & {\bf 304}\\
        \bottomrule
    \end{tabular}
    }
    \end{subtable}
    \begin{subtable}[t]{.3\linewidth}\centering
    \caption{\footnotesize\textbf{Clips and objects fusion}} 
    \label{tab:ablation_fusion}
    \scalebox{0.7}{
    \begin{tabular}{lcccc}
        \toprule
         & ~$K$=1 & ~$K$=10 & ~$K$=100 & ~MR $\downarrow$ \\
        \midrule
        Early & 0.59 & 3.77 & 20.78 & 506 \\
        Late-sep & {\bf 1.41} & 7.40 & 30.23 & {\bf 304}\\
        Late-joint & 1.38 & {\bf 7.92} & {\bf 30.56} & 309 \\
        \bottomrule
    \end{tabular}
    }
    \end{subtable}
    \\
    \vspace{6pt}
    \begin{subtable}[t]{.3\linewidth}\centering
    \caption{\footnotesize\textbf{Intra-negatives}} 
    \label{tab:ablation_intra_neg}
    \scalebox{0.7}{
    \begin{tabular}{lcccc}
        \toprule
        $|\negativesetintra|^{(p)}$ & ~$K$=1 & ~$K$=10 & ~$K$=100 & ~MR $\downarrow$ \\
        \midrule
        1 & {\bf 1.38} & {\bf 7.92} & {\bf 30.56} & {\bf 309} \\
        3 & 1.08 & 6.03 & 24.34 & 468 \\
        5 & 1.10 & 6.16 & 26.15 & 431 \\
        10 & 0.69 & 3.78 & 14.68 & 887 \\
        \bottomrule
    \end{tabular}
    }
    \end{subtable}
    \begin{subtable}[t]{.3\linewidth}\centering
    \caption{\footnotesize\textbf{Inter-negatives}} \label{tab:ablation_inter_neg}
    \scalebox{0.7}{
    \begin{tabular}{lcccc}
        \toprule
        $|\negativesetinter|^{(p)}$ & ~$K$=1 & ~$K$=10 & ~$K$=100 & ~MR $\downarrow$ \\
        \midrule
        1 & 1.38 & 7.92 & 30.56 & 309 \\
        5 & 1.60 & 8.61 & 33.06 & 270 \\
        10 & {\bf 1.69} & {\bf 9.26} & {\bf 34.52} & {\bf 249} \\
        50 & 1.58 & 9.06 & 34.77 & 258 \\
        \bottomrule
    \end{tabular}
    }
    \end{subtable}
    \begin{subtable}[t]{.3\linewidth}\centering
    \caption{\footnotesize\textbf{NCE vs.\ triplet loss}} \label{tab:ablation_loss}
    \scalebox{0.7}{
    \begin{tabular}{lcccc}
        \toprule
         & ~$K$=1 & ~$K$=10 & ~$K$=100 & ~MR $\downarrow$ \\
        \midrule
        Triplet & 1.69 & 9.26 & 34.52 & 249 \\
        NCE & {\bf 1.95} & {\bf 9.62} & {\bf 35.80} & {\bf 242} \\
        \bottomrule
    \end{tabular}
    }
  \end{subtable}

\vspace{-2pt}
\end{table*}

\subsection{Video corpus moment retrieval}
\label{sec:video_corpus}

Our first experiment consists of exhaustively evaluating a method over an entire video corpus. More specifically, given a language query, we evaluate the alignment cost exhaustively over all possible moments in all videos. We describe in detail our evaluation setup and results.

\paragraph{Baselines.} 
We compare our \modelnameabbr{} model to MCN~\cite{hendricks2017localizing}, a competitive baseline for the single video moment retrieval task, run exhaustively over all moments in the corpus. 
We re-implemented MCN~\cite{hendricks2017localizing} and train it using ResNet-152 features~\cite{He2016DeepRL} which significantly improves the performance of the single video retrieval task as reported in Hendricks \etal~\cite{hendricks2017localizing}.  
See the Appendix for details.
As a second baseline, we explore the use of action proposals obtained from R-C3D~\cite{Xu_2017_ICCV} instead of sliding-window candidate moments as used in MCN. For this baseline, we incorporate the action proposals inside the MCN model and evaluate in the exhaustive setting to determine their efficacy in the newly proposed task of natural language video corpus moment retrieval.
In addition, we compare to chance, moment-frequency-prior (``Moment prior''), and temporal endpoint (``TEF-only'') baselines. 
For chance, we return moments across all videos sampled from a uniform distribution. 
We compute the moment frequency prior as in Hendricks \etal~\cite{hendricks2017localizing} for each dataset by discretizing the range of video-length-normalized start and end points and histograming the training ground truth moments. 
We output the probability for each video's moment; ties across different videos are broken by sampling a uniform distribution. 
We train the MCN model using the same procedure as for single video moment retrieval~\cite{hendricks2017localizing}. 
For TEF-only, we train the MCN~\cite{hendricks2017localizing} model using only the language features and TEF on all the datasets, \ie, the model does not see appearance features from the video.

\begin{table*}[!t]
\caption{
\label{tab:corpus_quantitative}
	{\bf \footnotesize Video corpus retrieval quantitative results (exhaustive setting).} 
	\footnotesize We show average recall for top $K$ retrievals and median retrieval rank (MR, lower is better) averaged over IoU=$\{0.5,0.7\}$ on DiDeMo~\cite{hendricks2017localizing} dataset for different baselines and our model. More details in text.
}
\normalsize
\centering 
\footnotesize
\resizebox{0.6\linewidth}{!}{%
\begin{tabular}{lc | cccc | cccc }
\toprule 

& &
\multicolumn{4}{c|}{DiDeMo~\cite{hendricks2017localizing} (test)}  &
\multicolumn{4}{c}{Charades-STA~\cite{gao2017tall} (test)}
\\

& TEF &
~$K$=1 & ~$K$=10 & ~$K$=100 & ~MR $\downarrow$ &
~$K$=1 & ~$K$=10 & ~$K$=100 & ~MR $\downarrow$ \\
\midrule

 Chance & \xmark &
0.00 & 0.06 & 1.32 & 8834 &
0.01 & 0.06 & 0.74 & 11732 \\

 Moment prior & \xmark &
0.02 & 0.20 & 2.17 & 2881 &
0.02 & 0.11 & 1.10 & 8303 \\

 TEF-only & \cmark &
0.04 & 0.25 & 2.19 & 2906 &
0.03 & 0.25 & 2.22 & 6273 \\

\midrule

 MCN & \xmark &
0.30 & 2.20 & 11.44 & 1181 &
0.13 & 0.82 & 3.84 & 7598 \\

STAL (clips) & \xmark &
1.55 & 6.40 & 22.18 & 536 &
0.19 & 1.25 & 4.30 & 7055 \\

\midrule

 MCN & \cmark &
1.03 & 5.76 & 26.67 & 354 &
0.23 & 1.24 & 5.55 & 3902 \\

MCN (R-C3D) & \cmark &
- & - & - & - &
0.15 & 0.36 & 3.88 & 5099 \\

 Ours & \cmark &
{\bf 2.25} & {\bf 10.40} & {\bf 36.46} & {\bf 234} &
{\bf 0.31} & {\bf 2.02} & {\bf 9.80} & {\bf 2751} \\

\bottomrule 
\end{tabular}
}
\vspace{-5pt}
\end{table*}

\begin{figure*}[t]
	\centering
	\includegraphics[width=0.98\textwidth]{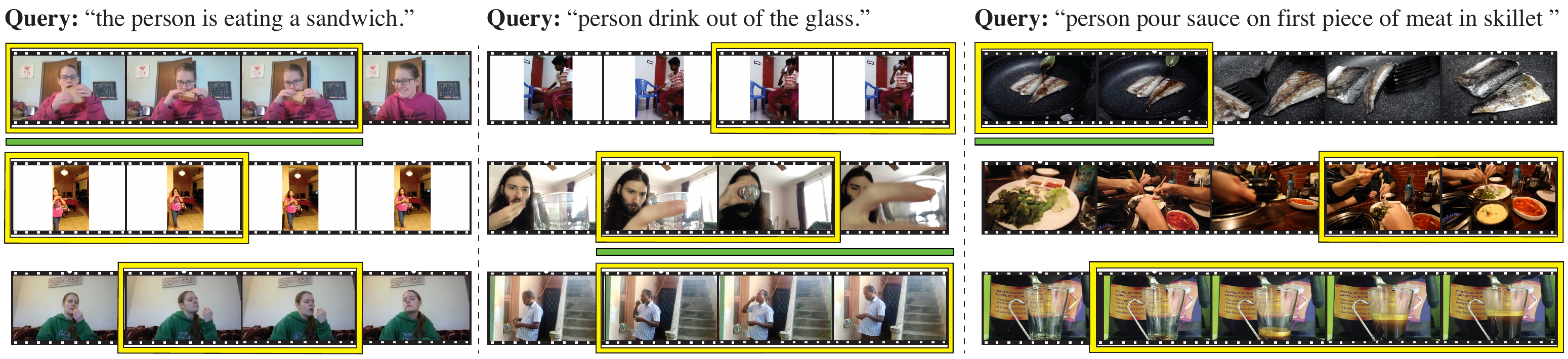}
	\caption{
	{\bf \footnotesize Video corpus retrieval qualitative results.} \footnotesize We show top temporally localized moment retrievals for different natural langauge queries across all videos in DiDeMo~\cite{hendricks2017localizing} and Charades-STA~\cite{gao2017tall}. 
	Ground truth annotations appear as a green line below a video, best viewed in color. 
	Refer to the appendix for videos and more results.
	}
	\label{fig:corpus_qualitative}
\end{figure*}

\paragraph{Ablation study.}
We justify all the design choices of our proposed model on the validation set of DiDeMo in the exhaustive setting. All models use TEF features. 

\noindent{\it Clip vs.\ moment alignment}. 
First, we seek to evaluate the effectiveness of clip-level alignment for our proposed task compared to moment alignment. 
We evaluate our STAL (clips) approach against MCN and report results in 
Table~\ref{tab:ablation_alignment}. Our Clips-only alignment out-performs MCN over all the evaluation criteria, demonstrating the effectiveness of fine-grained alignment at the clip level. 

\noindent{\it Spatiotemporal alignment}. 
Next, we seek to evaluate the effectiveness of spatiotemporal alignment. 
We compare spatiotemporal video features consisting of clips only (``Clips''), detected objects only (``Objects''), and combined feature set of clips and detected objects (``Clips + objects'') and report results in Table~\ref{tab:ablation_spatio_temporal}. 
We observed that objects-only does not perform well. However, when objects are combined with clips, we observe the best accuracy, thus validating our spatiotemporal alignment approach.

\noindent{\it Clip and objects fusion}.~
Next, we evaluate different strategies for fusing the clips and objects features. 
The strategies include concatenating the clips and detected objects features (``Early''), performing alignment over the two sets separately, where networks for each set is trained separately and the final score is the convex combination of the two scores (``Late-sep''), and performing separate clip/object alignment but training the two branches jointly (``Late-joint'').
We report results for the different strategies in Table~\ref{tab:ablation_fusion}. 
We observe that late-fusion significantly outperforms early-fusion on all criteria.
Meanwhile, joint training gives a slight advantage over separate models for $K\in\{10,100\}$. We use joint training for the rest of the experiments due to the practical benefits of end-to- training.

\noindent{\it Impact of intra- and inter-negatives}.
We study the impact of varying the number of intra- and inter-negatives samples per moment during training.
We report results in Tables~\ref{tab:ablation_intra_neg} and~\ref{tab:ablation_inter_neg}. We conclude that increasing the number of intra negatives impairs the performance; conversely, increasing the number of inter negatives is beneficial. For the latter, we observe that performance saturates around 10 negatives per candidate moment.

\noindent{\it NCE vs.\ triplet loss}.
We seek to understand the relevance of the NCE training objective. Up until here, we have used the triplet loss~\cite{chopraHL2005_ranking,schultzT2003_triplet} for fair comparison with~\cite{hendricks2017localizing}. In particular, our triplet loss is the sum of ranking losses,
\begin{equation}
    \tripletloss_\modelparameters = 
    \sum_{p\in\trainingset}
    \sum_{n\in\allnegatives^{(p)}}
    \rankingloss{\left(\costalignmenttrain_p,\costalignmenttrain_n\right)}
    \label{eqn:trainingloss}
\end{equation}
where $\rankingloss\mathopen{}\left(x,y\right)\mathclose{} = \max\left(0,x-y+b\right)$ is a ranking loss and $b$ is the margin hyperparameter.
We set parameters $b=0.1$ using cross validation.
Table~\ref{tab:ablation_loss} showcases the impact of the NCE loss compared to the triplet loss.
We observe that the NCE loss out-performs the triplet loss for this task. Thus, we use NCE loss in the rest of this section.

\paragraph{Results.}
Quantitative results and comparisons to baselines are shown in Table~\ref{tab:corpus_quantitative}. 
Our \modelnameabbr{} model out-performs all baselines in recall across both datasets, validating the effectiveness of our approach on our newly proposed task.

In particular, we obtain a $37\%-118\%$ and up to $30\%$ relative improvement over MCN with TEF for average recall and median rank, respectively. 
We note that the performance is low for all methods as annotations are not exhaustive and there are significantly more candidate moments to search over than in the single video retrieval task, illustrating the difficulty of the video corpus retrieval task.
We hypothesize that the lower MTLD of the Charade-STA dataset (\cf, Table~\ref{tab:dataset_settings}) could explain why the performance is lower for this dataset. A low lexical diversity can hamper the generation of discriminative query features, making the ranking operation harder.

\begin{table*}[t]
\caption{
\label{tab:corpus_quantitative_twostage}
	{\bf \footnotesize  Efficient retrieval with re-ranking quantitative results.} 
	\footnotesize  We show average recall for top $K$ retrievals averaged over IoU=$\{0.5,0.7\}$ on DiDeMo~\cite{hendricks2017localizing}, Charades-STA~\cite{gao2017tall}, for different baselines and our model. We also report run-time (s) for the retrieval and re-ranking stages (retrieval / re-ranking), and index size (GB) in a hypothetical 1M video collection, modelled using random feature vectors. More details in text. 
}
\centering 
\footnotesize
\begin{tabular}{ll | ccc | ccc | cc}
\toprule

\multicolumn{1}{c}{\multirow{2}{*}{Retrieval}} & \multicolumn{1}{c|}{\multirow{2}{*}{Re-ranking}} & \multicolumn{3}{c|}{\multirow{2}{*}{DiDeMo (test)}} &
\multicolumn{3}{c|}{\multirow{2}{*}{Charades-STA (test)}} &
\multicolumn{2}{c}{1M Videos Copus}
\\
& & & & & & & & Run & Index 
\\
 \multicolumn{1}{c}{stage} & \multicolumn{1}{c|}{stage} & 
 ~$K$=1~ & ~$K$=10~ & ~$K$=100~ &
 ~$K$=1~ & ~$K$=10~ & $K$=100 &
 time (s) & size (GB)\\
\midrule

 MEE & MCN (TEF) &
0.48 & 2.55 & 6.40 &
0.22 & 	0.75 & 	2.00 &
~0.3 / 0.2 & 0.37 \\

 MEE & \cite{Xu19} &
 - & - & - &
  0.12 & 0.30 & 1.01 &
0.3 / ~-~ & 0.37 \\
 
MEE & CBP~\cite{WangMJ2020} &
 - & - & - &
0.10 & 0.43 & 1.79 &
0.3 / ~-~ & 0.37 \\

MEE & 2D-TAN~\cite{zhang2020learning} &
 - & - & - &
0.13 & 0.85 & 2.00 &
0.3 / ~-~ & 0.37 \\

 MCN & MCN (TEF) &
0.90 & 4.18 & 9.50 & 
0.22 & 0.99 & 2.10 &
136.3 / 0.2~~  & 63.3 \\

 \modelnameabbr{} (clips) & MCN (TEF) &
1.47 & 7.85 & 23.05&
0.23 & 1.08 & 3.52 &
24.0 / 0.2~ & 7.45 \\

 \modelnameabbr{} (clips) & \modelnameabbr{} (TEF) &
2.10 & 9.79 & 23.20 & 
0.36 & 1.41 & 2.81 &
24.0 / 0.4~  & 7.45 \\

 Approx.\ \modelnameabbr{} & \modelnameabbr{} (TEF) &
{\bf 2.14} & {\bf 10.27} & {\bf 24.38}&
{\bf 0.38} & {\bf 1.57} & {\bf 3.92} &
1.0 / 0.4  & 7.45 \\

\bottomrule 
\end{tabular}
\vspace{-5pt}
\end{table*}

Qualitative results are shown in Figure~\ref{fig:corpus_qualitative}. 
Notice how we are able to retrieve relevant moments for the different language queries. For example, the queries ``\textit{the person is eating a sandwich}'' and ``\textit{person drink out of the glass}'' retrieves well-localized moments depicting people eating or drinking, including single ground truth annotated moments for the queries. 
The query ``\textit{person pour sauce on first piece of meat in skillet}'' shows example failures of our system. While the top retrieval is correct, the other retrievals depict different parts of the language query, such as ``sauce'', ``meat'', and ``pour'', but not the entire query.

\subsection{Efficient retrieval with re-ranking}
\label{sec:video_corpus_twostage}

For our second experiment, we evaluate the efficient retrieval and re-ranking system described in Section~\ref{sec:approach}. 
In the {\em retrieval stage}, we retrieve the top 200 moments and successively apply the {\em re-ranking stage}, which re-arranges the ranking of the moments according to a more complex model.
We thoroughly investigate several combinations of different {\em retrieval and re-ranking} methods.

\paragraph{Evaluation criteria.} 
Similar to the exhaustive retrieval setting (Section~\ref{sec:video_corpus}), we report average recall at $K\in\left\{1, 10, 100\right\}$ on the DiDeMo and Charades-STA datasets for the video corpus moment retrieval task, averaged over IoU$\in\{0.5,0.7\}$. 
Note that we do not report median rank as only the top retrieved moments are considered. 

\paragraph{Baselines and ablations.} 
We use MCN or \modelnameabbr{} (clips) for the retrieval stage, followed by MCN, or \modelnameabbr{} with TEF for the re-ranking stage.
Note that \modelnameabbr{} (clips) does not rely on TEF in order to be agnostic to moment lengths  and effectively achieving a reduction of the index size (\cf, Section~\ref{sec:reranking}). 
We also consider using MEE~\cite{Miech2018} for the retrieval stage as it performs well on the LSMDC benchmark~\cite{Rohrbach2015ADF} and outperforms other recent methods~\cite{Dong18} on MSR-VTT~\cite{Xu2016MSRVTTAL} (\cf, supplemental).
We used the publicly available implementation of MEE to retrieve videos from the corpus and turned off flow, face, and audio features in MEE for fair comparison. 
We tried MEE pre-trained on LSMDC and MSR-VTT, which performed near chance on our task; retraining MEE on the target datasets performed best.
Along the same lines, we also use the approaches~\cite{WangMJ2020,Xu19,zhang2020learning} for the re-ranking stage to gauge the impact of other single video moment retrieval algorithms on our task.
These approaches perform considerably well on Charades-STA, and their implementation and pre-trained models are publicly available.
During retrieval with MEE, we maintain a comparable number of moments for the re-ranking stage by retrieving the top videos such that there are at least 200 available moments within the retrieved videos.
Finally, we consider the approximate retrieval setting where we retrieve the top 200 clips given a language query and consider moments around the retrieved clips for the re-ranking stage (Approx.\ \modelnameabbr{}).

\paragraph{Results.} 
We report quantitative results in Table~\ref{tab:corpus_quantitative_twostage}. 
We show how using \modelnameabbr{} (TEF) for re-ranking outperforms MCN (TEF) across all criteria in both datasets, demonstrating the effectiveness of our approach for retrieval with re-ranking. 
Moreover, our two-stage approach is on par or outperforms the exhaustive approaches in Table~\ref{tab:corpus_quantitative} for $K\in\left\{1, 10\right\}$; we achieve these results at a lower computational cost. 
We find that \modelnameabbr{} (clips) outperforms both MCN and MEE for first-stage retrieval when using the same method for the re-ranking stage. 
This result demonstrates the importance of indexing moment clips as opposed to a coarse aggregated video feature for our task. 
Similarly, we observe that for the same retrieval stage, MEE, the MCN (TEF) model  out-performs Text-to-Clip~\cite{Xu19}, CBP~\cite{WangXWQLTG2016} and 2D-TAN~\cite{zhang2020learning} approaches for all values of $K$. These results motivate our choice for directly comparing our approach against MCN (TEF) for re-ranking.
Our takeaway is that the standard single video moment retrieval methods have not been specifically trained for our new task (i.e., they do not leverage inter-video negatives during training) and therefore fail to generalize to the corpus setup. We conclude that these methods cannot be simply adopted off-the-shelf for retrieval at scale given their inefficiency and lower retrieval accuracy.
Finally, for approximate retrieval, we preserve high recall for $K\in\left\{1, 10\right\}$, demonstrating its effectiveness in an efficient retrieval setting.

\paragraph{Run time and index size.} 
We report the run time, in seconds, to process a query and the index size, in GB, of each approach in Table~\ref{tab:corpus_quantitative_twostage}. For fostering the development of scalable models for our task, we consider a hypothetical corpus containing 1M videos, which differs from the DiDeMo and Charades-STA datasets. Each video contains $20$ clips with max-moment length of $14$ clips for the different methods, which corresponds to a realistic scenario where a user queries a large corpus of videos. 
According to these parameters, we generate the indexing table (i.e., vector embeddings are generated by random sampling according to a uniform distribution over $[0,1)$)
for all methods following the specific strategy (indexing videos, moments, or clips) and measure the memory footprint (index size). Moreover, we perform $100$ repeated queries for $100$ randomly generated query embeddings and average the clock time to estimate the run-time. Note that the indexing strategy (videos, moments, or clips) determines the size of the indexing table and, therefore, impacts both the memory footprint and run-time estimation. Conversely, the search method (exhaustive/approximate) only impacts the run-time. The experiments are conducted on an Intel Xeon(R) Gold $6240$R CPU @ $2.40$GHz$\times 48$, no GPU acceleration is used and random generated numbers are of type float32.
We report the timings for the two stages separately (retrieval/re-ranking) in Table~\ref{tab:corpus_quantitative_twostage}.

First, observe that MCN is the slowest method for the first-stage retrieval, which is due to the large search space as the model indexes every possible proposal in the dataset.
In contrast, MEE achieves the fastest query time and the smallest index size, as the search space consists of only the dataset's videos, which is less than the number of clips and the number of moments. However, MEE is also characterized by the lowest accuracy regardless of the method used for the re-ranking stage, making it unsuitable for this task.
We argue that the low ranking accuracy is a result of entire video retrieval not having the granularity to identify all (potentially short) moments of interest contained in it. Conversely, we find that retrieving clips (instead of entire videos) in the first stage, although more expensive, yields a better trade-off between recall performance and space/time complexity.
STAL (clips) achieves a 5x reduction of the run time and 8x reduction in memory footprint compared to MCN while retaining better performance. 
The best trade-off between retrieval accuracy, run-time, and index size is obtained by Approx.\ STAL, which is 136x times faster and utilizes 8x less memory than MCN. 

For the re-ranking stage, we notice that MCN (TEF) is more efficient than STAL (TEF). The slower run time of STAL (TEF) is due to the use of object information and more expensive cost function.
For a retrieval stage based on STAL (clips), we notice that MCN (TEF) obtains competitive performance compared to STAL (TEF) in terms of recall $K=100$. However, STAL (TEF) retains better performance on the stringent criteria $K\in{1,10}$ at a negligible extra computational cost. The trade-off between performance and run-time is, therefore, more favourable for STAL (TEF), making it the most successful model for the re-ranking stage.

\subsection{Future directions}

Transformer-based architectures have shown promising results and could be adopted for our newly proposed task. Nonetheless, attention must be paid to the efficiency aspect to enable fast moment retrieval from the video corpus. First, joint vision-language transformers such as~\cite{lu2019vilbert,su2019vl,sun2019videobert}, which have shown promising results in several vision-language tasks, are not indexable. In contrast, dual encoder architectures such as CLIP~\cite{radford2021learning} or ALIGN~\cite{jia2021scaling} have demonstrated strong text-to-image retrieval performance when trained on very large datasets of hundreds of millions of images. Dual encoder methods are similar in spirit to our efficient STAL (Clips) and Approx STAL models in that they are amenable to retrieval using efficient indexing techniques. Yet, it remains to be seen if these dual encoder methods would scale well to large-scale video datasets. Early work in this direction~\cite{bain2021frozen,Luo2021CLIP4Clip,miech2021thinking} is promising. In addition to the compute issue, the aforementioned joint vision-language transformers address different tasks. \cite{lu2019vilbert,su2019vl} address language-based retrieval for still images. \cite{sun2019videobert} considers the video captioning and action classification tasks. The extension of the above work to retrieve variable length moments from a corpus, as we do in our work, is non-trivial and is an interesting direction for future work.

\section{Conclusion}

We have shown a simple yet effective approach for aligning video clips to natural language queries for retrieving moments in untrimmed, unsegmented videos. 
Our approach allows for efficient indexing and retrieval of video moments on our newly proposed task of search through large video collections. 
We have quantitatively evaluated on benchmark datasets extended to our task and shown the effectiveness of our approach over prior work on our proposed task in terms of accuracy and search index size. 
Our work opens up the possibility of effectively searching video at large scale with natural language interfaces. \\

\noindent\textbf{Acknowledgments.} This work was supported by the King Abdullah University of Science and Technology (KAUST) Office of Sponsored Research through the Visual Computing Center (VCC) funding.

\bibliographystyle{model2-names}
\bibliography{refs}

\begin{thebibliography}{81}
\expandafter\ifx\csname natexlab\endcsname\relax\def\natexlab#1{#1}\fi
\providecommand{\url}[1]{\texttt{#1}}
\providecommand{\href}[2]{#2}
\providecommand{\path}[1]{#1}
\providecommand{\DOIprefix}{doi:}
\providecommand{\ArXivprefix}{arXiv:}
\providecommand{\URLprefix}{URL: }
\providecommand{\Pubmedprefix}{pmid:}
\providecommand{\doi}[1]{\href{http://dx.doi.org/#1}{\path{#1}}}
\providecommand{\Pubmed}[1]{\href{pmid:#1}{\path{#1}}}
\providecommand{\bibinfo}[2]{#2}
\ifx\xfnm\relax \def\xfnm[#1]{\unskip,\space#1}\fi
\bibitem[{Alwassel et~al.(2018)Alwassel, Heilbron, Escorcia and
  Ghanem}]{Alwassel18}
\bibinfo{author}{Alwassel, H.}, \bibinfo{author}{Heilbron, F.C.},
  \bibinfo{author}{Escorcia, V.}, \bibinfo{author}{Ghanem, B.},
  \bibinfo{year}{2018}.
\newblock \bibinfo{title}{Diagnosing error in temporal action detectors}, in:
  \bibinfo{booktitle}{Proceedings of the European Conference on Computer Vision
  (ECCV)}.
\bibitem[{Anderson et~al.(2018)Anderson, He, Buehler, Teney, Johnson, Gould and
  Zhang}]{Anderson2017up-down}
\bibinfo{author}{Anderson, P.}, \bibinfo{author}{He, X.},
  \bibinfo{author}{Buehler, C.}, \bibinfo{author}{Teney, D.},
  \bibinfo{author}{Johnson, M.}, \bibinfo{author}{Gould, S.},
  \bibinfo{author}{Zhang, L.}, \bibinfo{year}{2018}.
\newblock \bibinfo{title}{Bottom-up and top-down attention for image captioning
  and visual question answering}, in: \bibinfo{booktitle}{Proceedings of the
  IEEE/CVF Conference on Computer Vision and Pattern Recognition (CVPR)}.
\bibitem[{Bain et~al.(2021)Bain, Nagrani, Varol and Zisserman}]{bain2021frozen}
\bibinfo{author}{Bain, M.}, \bibinfo{author}{Nagrani, A.},
  \bibinfo{author}{Varol, G.}, \bibinfo{author}{Zisserman, A.},
  \bibinfo{year}{2021}.
\newblock \bibinfo{title}{Frozen in time: A joint video and image encoder for
  end-to-end retrieval}.
\newblock \bibinfo{journal}{arXiv preprint arXiv:2104.00650} .
\bibitem[{Barrow et~al.(1977)Barrow, Tenenbaum, Bolles and Wolf}]{Barrow77}
\bibinfo{author}{Barrow, H.}, \bibinfo{author}{Tenenbaum, J.},
  \bibinfo{author}{Bolles, R.}, \bibinfo{author}{Wolf, H.},
  \bibinfo{year}{1977}.
\newblock \bibinfo{title}{Parametric correspondence and chamfer matching: two
  new techniques for image matching}, in: \bibinfo{booktitle}{5th International
  Joint Conference on Articial Intelligence}.
\bibitem[{Bojanowski et~al.(2015)Bojanowski, Lagugie, Grave, Bach, Laptev,
  Ponce and Schmid}]{Bojanowski2015WeaklySupervisedAO}
\bibinfo{author}{Bojanowski, P.}, \bibinfo{author}{Lagugie, R.},
  \bibinfo{author}{Grave, E.}, \bibinfo{author}{Bach, F.R.},
  \bibinfo{author}{Laptev, I.}, \bibinfo{author}{Ponce, J.},
  \bibinfo{author}{Schmid, C.}, \bibinfo{year}{2015}.
\newblock \bibinfo{title}{Weakly-supervised alignment of video with text}, in:
  \bibinfo{booktitle}{Proceedings of the IEEE International Conference on
  Computer Vision (ICCV)}.
\bibitem[{Chen et~al.(2018)Chen, Chen, Ma, Jie and Chua}]{Chen18}
\bibinfo{author}{Chen, J.}, \bibinfo{author}{Chen, X.}, \bibinfo{author}{Ma,
  L.}, \bibinfo{author}{Jie, Z.}, \bibinfo{author}{Chua, T.S.},
  \bibinfo{year}{2018}.
\newblock \bibinfo{title}{Temporally grounding natural sentence in video}, in:
  \bibinfo{booktitle}{Proceedings of the conference on Empirical Methods in
  Natural Language Processing (EMNLP)}.
\bibitem[{Chen et~al.(2019)Chen, Ma, Chen, Jie and Luo}]{ChenMCJL19}
\bibinfo{author}{Chen, J.}, \bibinfo{author}{Ma, L.}, \bibinfo{author}{Chen,
  X.}, \bibinfo{author}{Jie, Z.}, \bibinfo{author}{Luo, J.},
  \bibinfo{year}{2019}.
\newblock \bibinfo{title}{Localizing natural language in videos}, in:
  \bibinfo{booktitle}{Proceedings of the AAAI Conference on Artificial
  Intelligence}.
\bibitem[{Chen and Jiang(2019)}]{ChenJ2019}
\bibinfo{author}{Chen, S.}, \bibinfo{author}{Jiang, Y.G.},
  \bibinfo{year}{2019}.
\newblock \bibinfo{title}{Semantic proposal for activity localization in videos
  via sentence query}, in: \bibinfo{booktitle}{Proceedings of the AAAI
  Conference on Artificial Intelligence}.
\bibitem[{{Chen Shaoxiang, Jiang Yu-Gang}(2020)}]{chenhierarchical}
\bibinfo{author}{{Chen Shaoxiang, Jiang Yu-Gang}}, \bibinfo{year}{2020}.
\newblock \bibinfo{title}{{Hierarchical Visual-Textual Graph for Temporal
  Activity Localization via Language}}, in: \bibinfo{booktitle}{Proceedings of
  the European Conference on Computer Vision (ECCV)}.
\bibitem[{{Chen Zhenfang, Ma Lin, Luo Wenhan, Wong Kwan-Yee
  Kenneth}(2019)}]{chen-etal-2019-weakly}
\bibinfo{author}{{Chen Zhenfang, Ma Lin, Luo Wenhan, Wong Kwan-Yee Kenneth}},
  \bibinfo{year}{2019}.
\newblock \bibinfo{title}{{Weakly-Supervised Spatio-Temporally Grounding
  Natural Sentence in Video}}, in: \bibinfo{booktitle}{Proceedings of the 57th
  Annual Meeting of the Association for Computational Linguistics (ACL)}.
\bibitem[{Chopra et~al.(2005)Chopra, Hadsell and LeCun}]{chopraHL2005_ranking}
\bibinfo{author}{Chopra, S.}, \bibinfo{author}{Hadsell, R.},
  \bibinfo{author}{LeCun, Y.}, \bibinfo{year}{2005}.
\newblock \bibinfo{title}{Learning a similarity metric discriminatively, with
  application to face verification}, in: \bibinfo{booktitle}{Proceedings of the
  IEEE/CVF Conference on Computer Vision and Pattern Recognition (CVPR)}.
\bibitem[{Dong et~al.(2019)Dong, Li, Xu, Ji and Wang}]{Dong18}
\bibinfo{author}{Dong, J.}, \bibinfo{author}{Li, X.}, \bibinfo{author}{Xu, C.},
  \bibinfo{author}{Ji, S.}, \bibinfo{author}{Wang, X.}, \bibinfo{year}{2019}.
\newblock \bibinfo{title}{Dual dense encoding for zero-example video
  retrieval}, in: \bibinfo{booktitle}{Proceedings of the IEEE/CVF Conference on
  Computer Vision and Pattern Recognition (CVPR)}.
\bibitem[{Escorcia et~al.(2016)Escorcia, Heilbron, Niebles and
  Ghanem}]{EscorciaHNG16}
\bibinfo{author}{Escorcia, V.}, \bibinfo{author}{Heilbron, F.C.},
  \bibinfo{author}{Niebles, J.C.}, \bibinfo{author}{Ghanem, B.},
  \bibinfo{year}{2016}.
\newblock \bibinfo{title}{Daps: Deep action proposals for action
  understanding}, in: \bibinfo{booktitle}{Proceedings of the European
  Conference on Computer Vision (ECCV)}.
\bibitem[{Gao et~al.(2018)Gao, Chen and Nevatia}]{Gao2018CTAPCT}
\bibinfo{author}{Gao, J.}, \bibinfo{author}{Chen, K.},
  \bibinfo{author}{Nevatia, R.}, \bibinfo{year}{2018}.
\newblock \bibinfo{title}{{CTAP}: Complementary temporal action proposal
  generation}, in: \bibinfo{booktitle}{Proceedings of the European Conference
  on Computer Vision (ECCV)}.
\bibitem[{Gao et~al.(2017)Gao, Sun, Yang and Nevatia}]{gao2017tall}
\bibinfo{author}{Gao, J.}, \bibinfo{author}{Sun, C.}, \bibinfo{author}{Yang,
  Z.}, \bibinfo{author}{Nevatia, R.}, \bibinfo{year}{2017}.
\newblock \bibinfo{title}{{TALL}: Temporal activity localization via language
  query}, in: \bibinfo{booktitle}{Proceedings of the IEEE International
  Conference on Computer Vision (ICCV)}.
\bibitem[{Gao et~al.(2019)Gao, Davis, Socher and Xiong}]{GaoDSX2019}
\bibinfo{author}{Gao, M.}, \bibinfo{author}{Davis, L.},
  \bibinfo{author}{Socher, R.}, \bibinfo{author}{Xiong, C.},
  \bibinfo{year}{2019}.
\newblock \bibinfo{title}{{WSLLN}:weakly supervised natural language
  localization networks}, in: \bibinfo{booktitle}{Proceedings of the 2019
  Conference on Empirical Methods in Natural Language Processing and the 9th
  International Joint Conference on Natural Language Processing
  (EMNLP-IJCNLP)}, \bibinfo{publisher}{Proceedings of the 57th Annual Meeting
  of the Association for Computational Linguistics (ACL)}.
\bibitem[{Ge et~al.(2014)Ge, He, Ke and Sun}]{Ge2014OptimizedPQ}
\bibinfo{author}{Ge, T.}, \bibinfo{author}{He, K.}, \bibinfo{author}{Ke, Q.},
  \bibinfo{author}{Sun, J.}, \bibinfo{year}{2014}.
\newblock \bibinfo{title}{Optimized product quantization}.
\newblock \bibinfo{journal}{IEEE Transactions on Pattern Analysis and Machine
  Intelligence} .
\bibitem[{Ghosh et~al.(2019)Ghosh, Agarwal, Parekh and
  Hauptmann}]{GhoshAPH2019}
\bibinfo{author}{Ghosh, S.}, \bibinfo{author}{Agarwal, A.},
  \bibinfo{author}{Parekh, Z.}, \bibinfo{author}{Hauptmann, A.},
  \bibinfo{year}{2019}.
\newblock \bibinfo{title}{Excl: Extractive clip localization using natural
  language descriptions}.
\newblock \bibinfo{journal}{Proceedings of the 57th Annual Meeting of the
  Association for Computational Linguistics (ACL)} .
\bibitem[{Goyal et~al.(2018)Goyal, Khot, Summers-Stay, Batra and
  Parikh}]{Goyal2018MakingTV}
\bibinfo{author}{Goyal, Y.}, \bibinfo{author}{Khot, T.},
  \bibinfo{author}{Summers-Stay, D.}, \bibinfo{author}{Batra, D.},
  \bibinfo{author}{Parikh, D.}, \bibinfo{year}{2018}.
\newblock \bibinfo{title}{Making the {V} in {VQA} matter: Elevating the role of
  image understanding in visual question answering}.
\newblock \bibinfo{journal}{International Journal of Computer Vision} .
\bibitem[{Gray and Neuhoff(1998)}]{Gray1998QuantizationI}
\bibinfo{author}{Gray, R.M.}, \bibinfo{author}{Neuhoff, D.L.},
  \bibinfo{year}{1998}.
\newblock \bibinfo{title}{Quantization}.
\newblock \bibinfo{journal}{IEEE Transactions on Information Theory} .
\bibitem[{Gutmann and Hyv{\"a}rinen(2010)}]{gutmann2010_nce}
\bibinfo{author}{Gutmann, M.}, \bibinfo{author}{Hyv{\"a}rinen, A.},
  \bibinfo{year}{2010}.
\newblock \bibinfo{title}{Noise-contrastive estimation: A new estimation
  principle for unnormalized statistical models}, in:
  \bibinfo{booktitle}{Proceedings of the International Conference on Artificial
  Intelligence and Statistics (AISTATS)}, pp. \bibinfo{pages}{297--304}.
\bibitem[{Harris et~al.(2020)Harris, Millman, van~der Walt, Gommers, Virtanen,
  Cournapeau, Wieser, Taylor, Berg, Smith, Kern, Picus, Hoyer, van Kerkwijk,
  Brett, Haldane, del R{'{\i}}o, Wiebe, Peterson, G{'{e}}rard-Marchant,
  Sheppard, Reddy, Weckesser, Abbasi, Gohlke and Oliphant}]{numpy}
\bibinfo{author}{Harris, C.R.}, \bibinfo{author}{Millman, K.J.},
  \bibinfo{author}{van~der Walt, S.J.}, \bibinfo{author}{Gommers, R.},
  \bibinfo{author}{Virtanen, P.}, \bibinfo{author}{Cournapeau, D.},
  \bibinfo{author}{Wieser, E.}, \bibinfo{author}{Taylor, J.},
  \bibinfo{author}{Berg, S.}, \bibinfo{author}{Smith, N.J.},
  \bibinfo{author}{Kern, R.}, \bibinfo{author}{Picus, M.},
  \bibinfo{author}{Hoyer, S.}, \bibinfo{author}{van Kerkwijk, M.H.},
  \bibinfo{author}{Brett, M.}, \bibinfo{author}{Haldane, A.},
  \bibinfo{author}{del R{'{\i}}o, J.F.}, \bibinfo{author}{Wiebe, M.},
  \bibinfo{author}{Peterson, P.}, \bibinfo{author}{G{'{e}}rard-Marchant, P.},
  \bibinfo{author}{Sheppard, K.}, \bibinfo{author}{Reddy, T.},
  \bibinfo{author}{Weckesser, W.}, \bibinfo{author}{Abbasi, H.},
  \bibinfo{author}{Gohlke, C.}, \bibinfo{author}{Oliphant, T.E.},
  \bibinfo{year}{2020}.
\newblock \bibinfo{title}{Array programming with {NumPy}}.
\newblock \bibinfo{journal}{Nature} \bibinfo{volume}{585},
  \bibinfo{pages}{357--362}.
\newblock \DOIprefix\doi{10.1038/s41586-020-2649-2}.
\bibitem[{He et~al.(2016)He, Zhang, Ren and Sun}]{He2016DeepRL}
\bibinfo{author}{He, K.}, \bibinfo{author}{Zhang, X.}, \bibinfo{author}{Ren,
  S.}, \bibinfo{author}{Sun, J.}, \bibinfo{year}{2016}.
\newblock \bibinfo{title}{Deep residual learning for image recognition}, in:
  \bibinfo{booktitle}{Proceedings of the IEEE/CVF Conference on Computer Vision
  and Pattern Recognition (CVPR)}.
\bibitem[{Hendricks et~al.(2017)Hendricks, Wang, Shechtman, Sivic, Darrell and
  Russell}]{hendricks2017localizing}
\bibinfo{author}{Hendricks, L.A.}, \bibinfo{author}{Wang, O.},
  \bibinfo{author}{Shechtman, E.}, \bibinfo{author}{Sivic, J.},
  \bibinfo{author}{Darrell, T.}, \bibinfo{author}{Russell, B.},
  \bibinfo{year}{2017}.
\newblock \bibinfo{title}{Localizing moments in video with natural language},
  in: \bibinfo{booktitle}{Proceedings of the IEEE International Conference on
  Computer Vision (ICCV)}.
\bibitem[{Hendricks et~al.(2018)Hendricks, Wang, Shechtman, Sivic, Darrell and
  Russell}]{hendricks2018emnlp}
\bibinfo{author}{Hendricks, L.A.}, \bibinfo{author}{Wang, O.},
  \bibinfo{author}{Shechtman, E.}, \bibinfo{author}{Sivic, J.},
  \bibinfo{author}{Darrell, T.}, \bibinfo{author}{Russell, B.},
  \bibinfo{year}{2018}.
\newblock \bibinfo{title}{Localizing moments in video with temporal language},
  in: \bibinfo{booktitle}{Proceedings of the conference on Empirical Methods in
  Natural Language Processing (EMNLP)}.
\bibitem[{J{\'e}gou et~al.(2011)J{\'e}gou, Douze and
  Schmid}]{Jgou2011ProductQF}
\bibinfo{author}{J{\'e}gou, H.}, \bibinfo{author}{Douze, M.},
  \bibinfo{author}{Schmid, C.}, \bibinfo{year}{2011}.
\newblock \bibinfo{title}{Product quantization for nearest neighbor search}.
\newblock \bibinfo{journal}{IEEE Transactions on Pattern Analysis and Machine
  Intelligence} .
\bibitem[{Jia et~al.(2021)Jia, Yang, Xia, Chen, Parekh, Pham, Le, Sung, Li and
  Duerig}]{jia2021scaling}
\bibinfo{author}{Jia, C.}, \bibinfo{author}{Yang, Y.}, \bibinfo{author}{Xia,
  Y.}, \bibinfo{author}{Chen, Y.T.}, \bibinfo{author}{Parekh, Z.},
  \bibinfo{author}{Pham, H.}, \bibinfo{author}{Le, Q.V.},
  \bibinfo{author}{Sung, Y.}, \bibinfo{author}{Li, Z.},
  \bibinfo{author}{Duerig, T.}, \bibinfo{year}{2021}.
\newblock \bibinfo{title}{Scaling up visual and vision-language representation
  learning with noisy text supervision}.
\newblock \bibinfo{journal}{arXiv preprint arXiv:2102.05918} .
\bibitem[{Johnson et~al.(2019)Johnson, Douze and
  J{\'e}gou}]{Johnson2017BillionscaleSS}
\bibinfo{author}{Johnson, J.}, \bibinfo{author}{Douze, M.},
  \bibinfo{author}{J{\'e}gou, H.}, \bibinfo{year}{2019}.
\newblock \bibinfo{title}{Billion-scale similarity search with gpus}.
\newblock \bibinfo{journal}{IEEE Transactions on Big Data} .
\bibitem[{Khoreva et~al.(2018)Khoreva, Rohrbach and Schiele}]{Khoreva18}
\bibinfo{author}{Khoreva, A.}, \bibinfo{author}{Rohrbach, A.},
  \bibinfo{author}{Schiele, B.}, \bibinfo{year}{2018}.
\newblock \bibinfo{title}{Video object segmentation with language referring
  expressions}, in: \bibinfo{booktitle}{ACCV}.
\bibitem[{Krishna et~al.(2017)Krishna, Hata, Ren, Fei-Fei and
  Niebles}]{krishna2017dense}
\bibinfo{author}{Krishna, R.}, \bibinfo{author}{Hata, K.},
  \bibinfo{author}{Ren, F.}, \bibinfo{author}{Fei-Fei, L.},
  \bibinfo{author}{Niebles, J.C.}, \bibinfo{year}{2017}.
\newblock \bibinfo{title}{Dense-captioning events in videos}, in:
  \bibinfo{booktitle}{Proceedings of the IEEE International Conference on
  Computer Vision (ICCV)}.
\bibitem[{Krishna et~al.(2016)Krishna, Zhu, Groth, Johnson, Hata, Kravitz,
  Chen, Kalantidis, Li, Shamma, Bernstein and Fei-Fei}]{krishnavisualgenome}
\bibinfo{author}{Krishna, R.}, \bibinfo{author}{Zhu, Y.},
  \bibinfo{author}{Groth, O.}, \bibinfo{author}{Johnson, J.},
  \bibinfo{author}{Hata, K.}, \bibinfo{author}{Kravitz, J.},
  \bibinfo{author}{Chen, S.}, \bibinfo{author}{Kalantidis, Y.},
  \bibinfo{author}{Li, L.J.}, \bibinfo{author}{Shamma, D.A.},
  \bibinfo{author}{Bernstein, M.}, \bibinfo{author}{Fei-Fei, L.},
  \bibinfo{year}{2016}.
\newblock \bibinfo{title}{Visual genome: Connecting language and vision using
  crowdsourced dense image annotations}, in: \bibinfo{booktitle}{Proceedings of
  the IEEE/CVF Conference on Computer Vision and Pattern Recognition (CVPR)}.
\bibitem[{Lei et~al.(2020)Lei, Yu, Berg and Bansal}]{lei2020tvr}
\bibinfo{author}{Lei, J.}, \bibinfo{author}{Yu, L.}, \bibinfo{author}{Berg,
  T.L.}, \bibinfo{author}{Bansal, M.}, \bibinfo{year}{2020}.
\newblock \bibinfo{title}{{TVR: A Large-Scale Dataset for Video-Subtitle Moment
  Retrieval}}, in: \bibinfo{booktitle}{Proceedings of the European Conference
  on Computer Vision (ECCV)}.
\bibitem[{Li et~al.(2020)Li, Chen, Cheng, Gan, Yu and Liu}]{li-etal-2020-hero}
\bibinfo{author}{Li, L.}, \bibinfo{author}{Chen, Y.C.}, \bibinfo{author}{Cheng,
  Y.}, \bibinfo{author}{Gan, Z.}, \bibinfo{author}{Yu, L.},
  \bibinfo{author}{Liu, J.}, \bibinfo{year}{2020}.
\newblock \bibinfo{title}{{HERO}: Hierarchical encoder for {V}ideo+{L}anguage
  omni-representation pre-training}, in: \bibinfo{booktitle}{Proceedings of the
  2020 Conference on Empirical Methods in Natural Language Processing (EMNLP)}.
\bibitem[{Liu et~al.(2018a)Liu, Yeung, Chou, Huang, Fei-Fei and
  Niebles}]{liu2018tmn}
\bibinfo{author}{Liu, B.}, \bibinfo{author}{Yeung, S.}, \bibinfo{author}{Chou,
  E.}, \bibinfo{author}{Huang, D.A.}, \bibinfo{author}{Fei-Fei, L.},
  \bibinfo{author}{Niebles, J.C.}, \bibinfo{year}{2018}a.
\newblock \bibinfo{title}{Temporal modular networks for retrieving complex
  compositional activities in videos}, in: \bibinfo{booktitle}{Proceedings of
  the European Conference on Computer Vision (ECCV)}.
\bibitem[{Liu et~al.(2020)Liu, Qu, Liu, Dong, Zhou and Xu}]{liu2020jointly}
\bibinfo{author}{Liu, D.}, \bibinfo{author}{Qu, X.}, \bibinfo{author}{Liu, X.},
  \bibinfo{author}{Dong, J.}, \bibinfo{author}{Zhou, P.}, \bibinfo{author}{Xu,
  Z.}, \bibinfo{year}{2020}.
\newblock \bibinfo{title}{Jointly cross- and self-modal graph attention network
  for query-based moment localization}, in: \bibinfo{booktitle}{Proceedings of
  the 28th ACM International Conference on Multimedia (MM’20)}.
\bibitem[{Liu et~al.(2018b)Liu, Wang, Nie, He, Chen and
  Chua}]{Liu2018AttentiveMR}
\bibinfo{author}{Liu, M.}, \bibinfo{author}{Wang, X.}, \bibinfo{author}{Nie,
  L.}, \bibinfo{author}{He, X.}, \bibinfo{author}{Chen, B.},
  \bibinfo{author}{Chua, T.S.}, \bibinfo{year}{2018}b.
\newblock \bibinfo{title}{Attentive moment retrieval in videos}, in:
  \bibinfo{booktitle}{The 41st International ACM SIGIR Conference on Research
  \& Development in Information Retrieval}.
\bibitem[{Liu et~al.(2019)Liu, Albanie, Nagrani and Zisserman}]{LiuANZ2019}
\bibinfo{author}{Liu, Y.}, \bibinfo{author}{Albanie, S.},
  \bibinfo{author}{Nagrani, A.}, \bibinfo{author}{Zisserman, A.},
  \bibinfo{year}{2019}.
\newblock \bibinfo{title}{Use what you have: Video retrieval using
  representations from collaborative experts}, in: \bibinfo{booktitle}{British
  Machine Vision Conference}.
\bibitem[{Lu et~al.(2019a)Lu, Chen, Tan, Li and Xiao}]{lu_etal_2019_debug}
\bibinfo{author}{Lu, C.}, \bibinfo{author}{Chen, L.}, \bibinfo{author}{Tan,
  C.}, \bibinfo{author}{Li, X.}, \bibinfo{author}{Xiao, J.},
  \bibinfo{year}{2019}a.
\newblock \bibinfo{title}{{DEBUG: A Dense Bottom-Up Grounding Approach for
  Natural Language Video Localization}}, in: \bibinfo{booktitle}{Proceedings of
  the 2019 Conference on Empirical Methods in Natural Language Processing and
  the 9th International Joint Conference on Natural Language Processing
  (EMNLP-IJCNLP)}.
\bibitem[{Lu et~al.(2019b)Lu, Batra, Parikh and Lee}]{lu2019vilbert}
\bibinfo{author}{Lu, J.}, \bibinfo{author}{Batra, D.}, \bibinfo{author}{Parikh,
  D.}, \bibinfo{author}{Lee, S.}, \bibinfo{year}{2019}b.
\newblock \bibinfo{title}{Vilbert: Pretraining task-agnostic visiolinguistic
  representations for vision-and-language tasks}.
\newblock \bibinfo{journal}{arXiv preprint arXiv:1908.02265} .
\bibitem[{Luo et~al.(2021)Luo, Ji, Zhong, Chen, Lei, Duan and
  Li}]{Luo2021CLIP4Clip}
\bibinfo{author}{Luo, H.}, \bibinfo{author}{Ji, L.}, \bibinfo{author}{Zhong,
  M.}, \bibinfo{author}{Chen, Y.}, \bibinfo{author}{Lei, W.},
  \bibinfo{author}{Duan, N.}, \bibinfo{author}{Li, T.}, \bibinfo{year}{2021}.
\newblock \bibinfo{title}{Clip4clip: An empirical study of clip for end to end
  video clip retrieval}.
\newblock \bibinfo{journal}{arXiv preprint arXiv:2104.08860} .
\bibitem[{{Ma Minuk, Yoon Sunjae, Kim Junyeong, Lee Youngjoon, Kang Sunghun,
  Yoo Chang}(2020)}]{VLANet_ECCV_20}
\bibinfo{author}{{Ma Minuk, Yoon Sunjae, Kim Junyeong, Lee Youngjoon, Kang
  Sunghun, Yoo Chang}}, \bibinfo{year}{2020}.
\newblock \bibinfo{title}{{VLANet: Video-Language Alignment Network for
  Weakly-Supervised Video Moment Retrieval}}, in:
  \bibinfo{booktitle}{Proceedings of the European Conference on Computer Vision
  (ECCV)}.
\bibitem[{McCarthy(2005)}]{mccarthy2005assessment}
\bibinfo{author}{McCarthy, P.M.}, \bibinfo{year}{2005}.
\newblock \bibinfo{title}{An assessment of the range and usefulness of lexical
  diversity measures and the potential of the measure of textual, lexical
  diversity (MTLD)}.
\newblock Ph.D. thesis. The University of Memphis.
\bibitem[{McCarthy and Jarvis(2010)}]{mccarthyJ2010_mtld}
\bibinfo{author}{McCarthy, P.M.}, \bibinfo{author}{Jarvis, S.},
  \bibinfo{year}{2010}.
\newblock \bibinfo{title}{Mtld, vocd-d, and hd-d: A validation study of
  sophisticated approaches to lexical diversity assessment}.
\newblock \bibinfo{journal}{Behavior research methods} \bibinfo{volume}{42},
  \bibinfo{pages}{381--392}.
\bibitem[{Miech et~al.(2017)Miech, Alayrac, Bojanowski, Laptev and
  Sivic}]{Miech2017LearningFV}
\bibinfo{author}{Miech, A.}, \bibinfo{author}{Alayrac, J.B.},
  \bibinfo{author}{Bojanowski, P.}, \bibinfo{author}{Laptev, I.},
  \bibinfo{author}{Sivic, J.}, \bibinfo{year}{2017}.
\newblock \bibinfo{title}{Learning from video and text via large-scale
  discriminative clustering}, in: \bibinfo{booktitle}{Proceedings of the IEEE
  International Conference on Computer Vision (ICCV)}.
\bibitem[{Miech et~al.(2021)Miech, Alayrac, Laptev, Sivic and
  Zisserman}]{miech2021thinking}
\bibinfo{author}{Miech, A.}, \bibinfo{author}{Alayrac, J.B.},
  \bibinfo{author}{Laptev, I.}, \bibinfo{author}{Sivic, J.},
  \bibinfo{author}{Zisserman, A.}, \bibinfo{year}{2021}.
\newblock \bibinfo{title}{Thinking fast and slow: Efficient text-to-visual
  retrieval with transformers}, in: \bibinfo{booktitle}{Proceedings of the
  IEEE/CVF Conference on Computer Vision and Pattern Recognition}, pp.
  \bibinfo{pages}{9826--9836}.
\bibitem[{Miech et~al.(2018)Miech, Laptev and Sivic}]{Miech2018}
\bibinfo{author}{Miech, A.}, \bibinfo{author}{Laptev, I.},
  \bibinfo{author}{Sivic, J.}, \bibinfo{year}{2018}.
\newblock \bibinfo{title}{Learning a text-video embedding from incomplete and
  heterogeneous data}.
\newblock \bibinfo{journal}{CoRR} \href{http://arxiv.org/abs/1804.02516}{\tt
  arXiv:1804.02516}.
\bibitem[{Miech et~al.(2019)Miech, Zhukov, Alayrac, Tapaswi, Laptev and
  Sivic}]{MiechZATLS2019}
\bibinfo{author}{Miech, A.}, \bibinfo{author}{Zhukov, D.},
  \bibinfo{author}{Alayrac, J.B.}, \bibinfo{author}{Tapaswi, M.},
  \bibinfo{author}{Laptev, I.}, \bibinfo{author}{Sivic, J.},
  \bibinfo{year}{2019}.
\newblock \bibinfo{title}{How{T}o100{M}: {L}earning a {T}ext-{V}ideo
  {E}mbedding by {W}atching {H}undred {M}illion {N}arrated {V}ideo {C}lips},
  in: \bibinfo{booktitle}{Proceedings of the IEEE International Conference on
  Computer Vision (ICCV)}.
\bibitem[{Mithun et~al.(2019)Mithun, Paul and Roy{-}Chowdhury}]{MithunPR19}
\bibinfo{author}{Mithun, N.C.}, \bibinfo{author}{Paul, S.},
  \bibinfo{author}{Roy{-}Chowdhury, A.K.}, \bibinfo{year}{2019}.
\newblock \bibinfo{title}{Weakly supervised video moment retrieval from text
  queries}, in: \bibinfo{booktitle}{Proceedings of the IEEE/CVF Conference on
  Computer Vision and Pattern Recognition (CVPR)}.
\bibitem[{Muja and Lowe(2009)}]{MujaL09}
\bibinfo{author}{Muja, M.}, \bibinfo{author}{Lowe, D.G.}, \bibinfo{year}{2009}.
\newblock \bibinfo{title}{Fast approximate nearest neighbors with automatic
  algorithm configuration}, in: \bibinfo{editor}{Ranchordas, A.},
  \bibinfo{editor}{Ara{\'{u}}jo, H.} (Eds.), \bibinfo{booktitle}{{VISAPP}}, pp.
  \bibinfo{pages}{331--340}.
\bibitem[{Mun et~al.(2020)Mun, Cho and Han}]{Mun_2020_CVPR}
\bibinfo{author}{Mun, J.}, \bibinfo{author}{Cho, M.}, \bibinfo{author}{Han,
  B.}, \bibinfo{year}{2020}.
\newblock \bibinfo{title}{{Local-Global Video-Text Interactions for Temporal
  Grounding}}, in: \bibinfo{booktitle}{Proceedings of the IEEE/CVF Conference
  on Computer Vision and Pattern Recognition (CVPR)}.
\bibitem[{van~den Oord et~al.(2018)van~den Oord, Li and
  Vinyals}]{VdoordLV2019_infonce}
\bibinfo{author}{van~den Oord, A.}, \bibinfo{author}{Li, Y.},
  \bibinfo{author}{Vinyals, O.}, \bibinfo{year}{2018}.
\newblock \bibinfo{title}{Representation learning with contrastive predictive
  coding}.
\newblock \bibinfo{journal}{CoRR} \bibinfo{volume}{abs/1807.03748}.
\newblock \URLprefix \url{http://arxiv.org/abs/1807.03748}.
\bibitem[{Otani et~al.(2020)Otani, Nakahima, Rahtu,  and
  Heikkila}]{otani2020challengesmr}
\bibinfo{author}{Otani, M.}, \bibinfo{author}{Nakahima, Y.},
  \bibinfo{author}{Rahtu, E.}, , \bibinfo{author}{Heikkila, J.},
  \bibinfo{year}{2020}.
\newblock \bibinfo{title}{Uncovering hidden challenges in query-based video
  moment retrieval}, in: \bibinfo{booktitle}{The British Machine Vision
  Conference (BMVC)}.
\bibitem[{Paszke et~al.(2019)Paszke, Gross, Massa, Lerer, Bradbury, Chanan,
  Killeen, Lin, Gimelshein, Antiga, Desmaison, Kopf, Yang, DeVito, Raison,
  Tejani, Chilamkurthy, Steiner, Fang, Bai and Chintala}]{pytorch}
\bibinfo{author}{Paszke, A.}, \bibinfo{author}{Gross, S.},
  \bibinfo{author}{Massa, F.}, \bibinfo{author}{Lerer, A.},
  \bibinfo{author}{Bradbury, J.}, \bibinfo{author}{Chanan, G.},
  \bibinfo{author}{Killeen, T.}, \bibinfo{author}{Lin, Z.},
  \bibinfo{author}{Gimelshein, N.}, \bibinfo{author}{Antiga, L.},
  \bibinfo{author}{Desmaison, A.}, \bibinfo{author}{Kopf, A.},
  \bibinfo{author}{Yang, E.}, \bibinfo{author}{DeVito, Z.},
  \bibinfo{author}{Raison, M.}, \bibinfo{author}{Tejani, A.},
  \bibinfo{author}{Chilamkurthy, S.}, \bibinfo{author}{Steiner, B.},
  \bibinfo{author}{Fang, L.}, \bibinfo{author}{Bai, J.},
  \bibinfo{author}{Chintala, S.}, \bibinfo{year}{2019}.
\newblock \bibinfo{title}{Pytorch: An imperative style, high-performance deep
  learning library}, in: \bibinfo{booktitle}{Advances in Neural Information
  Processing Systems ({NeurIPSW})}, pp. \bibinfo{pages}{8024--8035}.
\bibitem[{Peleg et~al.(1989)Peleg, Werman and Rom}]{Peleg89}
\bibinfo{author}{Peleg, S.}, \bibinfo{author}{Werman, M.},
  \bibinfo{author}{Rom, H.}, \bibinfo{year}{1989}.
\newblock \bibinfo{title}{A unified approach to the change of resolution: Space
  and gray-level}.
\newblock \bibinfo{journal}{{IEEE} Transactions on Pattern Analysis and Machine
  Intelligence} .
\bibitem[{Pennington et~al.(2014)Pennington, Socher and
  Manning}]{Pennington2014GloveGV}
\bibinfo{author}{Pennington, J.}, \bibinfo{author}{Socher, R.},
  \bibinfo{author}{Manning, C.D.}, \bibinfo{year}{2014}.
\newblock \bibinfo{title}{Glove: Global vectors for word representation}, in:
  \bibinfo{booktitle}{Proceedings of the conference on Empirical Methods in
  Natural Language Processing (EMNLP)}.
\bibitem[{Radford et~al.(2021)Radford, Kim, Hallacy, Ramesh, Goh, Agarwal,
  Sastry, Askell, Mishkin, Clark et~al.}]{radford2021learning}
\bibinfo{author}{Radford, A.}, \bibinfo{author}{Kim, J.W.},
  \bibinfo{author}{Hallacy, C.}, \bibinfo{author}{Ramesh, A.},
  \bibinfo{author}{Goh, G.}, \bibinfo{author}{Agarwal, S.},
  \bibinfo{author}{Sastry, G.}, \bibinfo{author}{Askell, A.},
  \bibinfo{author}{Mishkin, P.}, \bibinfo{author}{Clark, J.}, et~al.,
  \bibinfo{year}{2021}.
\newblock \bibinfo{title}{Learning transferable visual models from natural
  language supervision}.
\newblock \bibinfo{journal}{arXiv preprint arXiv:2103.00020} .
\bibitem[{Regneri et~al.(2013)Regneri, Rohrbach, Wetzel, Thater, Schiele and
  Pinkal}]{regneri2013grounding}
\bibinfo{author}{Regneri, M.}, \bibinfo{author}{Rohrbach, M.},
  \bibinfo{author}{Wetzel, D.}, \bibinfo{author}{Thater, S.},
  \bibinfo{author}{Schiele, B.}, \bibinfo{author}{Pinkal, M.},
  \bibinfo{year}{2013}.
\newblock \bibinfo{title}{Grounding action descriptions in videos}.
\newblock \bibinfo{journal}{Transactions of the Association for Computational
  Linguistics (TACL)} .
\bibitem[{Ren et~al.(2015)Ren, He, Girshick and Sun}]{renNIPS15fasterrcnn}
\bibinfo{author}{Ren, S.}, \bibinfo{author}{He, K.}, \bibinfo{author}{Girshick,
  R.}, \bibinfo{author}{Sun, J.}, \bibinfo{year}{2015}.
\newblock \bibinfo{title}{Faster {R-CNN}: Towards real-time object detection
  with region proposal networks}, in: \bibinfo{booktitle}{Advances in Neural
  Information Processing Systems ({NeurIPS})}.
\bibitem[{Rohrbach et~al.(2015)Rohrbach, Rohrbach, Tandon and
  Schiele}]{Rohrbach2015ADF}
\bibinfo{author}{Rohrbach, A.}, \bibinfo{author}{Rohrbach, M.},
  \bibinfo{author}{Tandon, N.}, \bibinfo{author}{Schiele, B.},
  \bibinfo{year}{2015}.
\newblock \bibinfo{title}{A dataset for movie description}, in:
  \bibinfo{booktitle}{Proceedings of the IEEE/CVF Conference on Computer Vision
  and Pattern Recognition (CVPR)}.
\bibitem[{Schultz and Joachims(2003)}]{schultzT2003_triplet}
\bibinfo{author}{Schultz, M.}, \bibinfo{author}{Joachims, T.},
  \bibinfo{year}{2003}.
\newblock \bibinfo{title}{Learning a distance metric from relative
  comparisons}, pp. \bibinfo{pages}{41--48}.
\bibitem[{Shao et~al.(2018)Shao, Xiong, Zhao, Huang, Qiao and Lin}]{Shao18}
\bibinfo{author}{Shao, D.}, \bibinfo{author}{Xiong, Y.}, \bibinfo{author}{Zhao,
  Y.}, \bibinfo{author}{Huang, Q.}, \bibinfo{author}{Qiao, Y.},
  \bibinfo{author}{Lin, D.}, \bibinfo{year}{2018}.
\newblock \bibinfo{title}{Find and focus: Retrieve and localize video events
  with natural language queries}, in: \bibinfo{booktitle}{Proceedings of the
  European Conference on Computer Vision (ECCV)}.
\bibitem[{Sigurdsson et~al.(2016)Sigurdsson, Varol, Wang, Farhadi, Laptev and
  Gupta}]{sigurdsson2016hollywood}
\bibinfo{author}{Sigurdsson, G.A.}, \bibinfo{author}{Varol, G.},
  \bibinfo{author}{Wang, X.}, \bibinfo{author}{Farhadi, A.},
  \bibinfo{author}{Laptev, I.}, \bibinfo{author}{Gupta, A.},
  \bibinfo{year}{2016}.
\newblock \bibinfo{title}{Hollywood in homes: Crowdsourcing data collection for
  activity understanding}, in: \bibinfo{booktitle}{Proceedings of the European
  Conference on Computer Vision (ECCV)}.
\bibitem[{Su et~al.(2019)Su, Zhu, Cao, Li, Lu, Wei and Dai}]{su2019vl}
\bibinfo{author}{Su, W.}, \bibinfo{author}{Zhu, X.}, \bibinfo{author}{Cao, Y.},
  \bibinfo{author}{Li, B.}, \bibinfo{author}{Lu, L.}, \bibinfo{author}{Wei,
  F.}, \bibinfo{author}{Dai, J.}, \bibinfo{year}{2019}.
\newblock \bibinfo{title}{Vl-bert: Pre-training of generic visual-linguistic
  representations}.
\newblock \bibinfo{journal}{arXiv preprint arXiv:1908.08530} .
\bibitem[{Sun et~al.(2019)Sun, Myers, Vondrick, Murphy and
  Schmid}]{sun2019videobert}
\bibinfo{author}{Sun, C.}, \bibinfo{author}{Myers, A.},
  \bibinfo{author}{Vondrick, C.}, \bibinfo{author}{Murphy, K.},
  \bibinfo{author}{Schmid, C.}, \bibinfo{year}{2019}.
\newblock \bibinfo{title}{Videobert: A joint model for video and language
  representation learning}, in: \bibinfo{booktitle}{Proceedings of the IEEE/CVF
  International Conference on Computer Vision}, pp.
  \bibinfo{pages}{7464--7473}.
\bibitem[{Wang et~al.(2020)Wang, Ma and Jiang}]{WangMJ2020}
\bibinfo{author}{Wang, J.}, \bibinfo{author}{Ma, L.}, \bibinfo{author}{Jiang,
  W.}, \bibinfo{year}{2020}.
\newblock \bibinfo{title}{Temporally grounding language queries in videos by
  contextual boundary-aware prediction}, in: \bibinfo{booktitle}{Proceedings of
  the AAAI Conference on Artificial Intelligence}.
\bibitem[{Wang et~al.(2016)Wang, Xiong, Wang, Qiao, Lin, Tang and {Val
  Gool}}]{WangXWQLTG2016}
\bibinfo{author}{Wang, L.}, \bibinfo{author}{Xiong, Y.}, \bibinfo{author}{Wang,
  Z.}, \bibinfo{author}{Qiao, Y.}, \bibinfo{author}{Lin, D.},
  \bibinfo{author}{Tang, X.}, \bibinfo{author}{{Val Gool}, L.},
  \bibinfo{year}{2016}.
\newblock \bibinfo{title}{Temporal segment networks: Towards good practices for
  deep action recognition}, in: \bibinfo{booktitle}{Proceedings of the European
  Conference on Computer Vision (ECCV)}.
\bibitem[{Wang et~al.(2019)Wang, Huang and Wang}]{WangHW2019}
\bibinfo{author}{Wang, W.}, \bibinfo{author}{Huang, Y.}, \bibinfo{author}{Wang,
  L.}, \bibinfo{year}{2019}.
\newblock \bibinfo{title}{Language-driven temporal activity localization: A
  semantic matching reinforcement learning model}, in:
  \bibinfo{booktitle}{Proceedings of the IEEE/CVF Conference on Computer Vision
  and Pattern Recognition (CVPR)}.
\bibitem[{Xu et~al.(2017)Xu, Das and Saenko}]{Xu_2017_ICCV}
\bibinfo{author}{Xu, H.}, \bibinfo{author}{Das, A.}, \bibinfo{author}{Saenko,
  K.}, \bibinfo{year}{2017}.
\newblock \bibinfo{title}{R-c3d: Region convolutional 3d network for temporal
  activity detection}, in: \bibinfo{booktitle}{Proceedings of the IEEE
  International Conference on Computer Vision (ICCV)}.
\bibitem[{Xu et~al.(2019)Xu, He, Plummer, Sigal, Sclaroff and Saenko}]{Xu19}
\bibinfo{author}{Xu, H.}, \bibinfo{author}{He, K.}, \bibinfo{author}{Plummer,
  B.A.}, \bibinfo{author}{Sigal, L.}, \bibinfo{author}{Sclaroff, S.},
  \bibinfo{author}{Saenko, K.}, \bibinfo{year}{2019}.
\newblock \bibinfo{title}{Multilevel language and vision integration for
  text-to-clip retrieval}, in: \bibinfo{booktitle}{Proceedings of the AAAI
  Conference on Artificial Intelligence}.
\bibitem[{Xu et~al.(2016)Xu, Mei, Yao and Rui}]{Xu2016MSRVTTAL}
\bibinfo{author}{Xu, J.}, \bibinfo{author}{Mei, T.}, \bibinfo{author}{Yao, T.},
  \bibinfo{author}{Rui, Y.}, \bibinfo{year}{2016}.
\newblock \bibinfo{title}{{MSR-VTT}: A large video description dataset for
  bridging video and language}, in: \bibinfo{booktitle}{Proceedings of the
  IEEE/CVF Conference on Computer Vision and Pattern Recognition (CVPR)}.
\bibitem[{{Yijun Song, Jingwen Wang, Lin Ma, Zhou Yu, Jun
  Yu}(2020)}]{song2020weaklysupervised}
\bibinfo{author}{{Yijun Song, Jingwen Wang, Lin Ma, Zhou Yu, Jun Yu}},
  \bibinfo{year}{2020}.
\newblock \bibinfo{title}{{Weakly-Supervised Multi-Level Attentional
  Reconstruction Network for Grounding Textual Queries in Videos}}.
\newblock \href{http://arxiv.org/abs/2003.07048}{\tt arXiv:2003.07048}.
\bibitem[{Yu et~al.(2018)Yu, Kim and Kim}]{Yu2018AJS}
\bibinfo{author}{Yu, Y.}, \bibinfo{author}{Kim, J.}, \bibinfo{author}{Kim, G.},
  \bibinfo{year}{2018}.
\newblock \bibinfo{title}{A joint sequence fusion model for video question
  answering and retrieval}, in: \bibinfo{booktitle}{Proceedings of the European
  Conference on Computer Vision (ECCV)}.
\bibitem[{Yu et~al.(2017)Yu, Ko, Choi and Kim}]{Yu2017EndtoEndCW}
\bibinfo{author}{Yu, Y.}, \bibinfo{author}{Ko, H.}, \bibinfo{author}{Choi, J.},
  \bibinfo{author}{Kim, G.}, \bibinfo{year}{2017}.
\newblock \bibinfo{title}{End-to-end concept word detection for video
  captioning, retrieval, and question answering}, in:
  \bibinfo{booktitle}{Proceedings of the IEEE/CVF Conference on Computer Vision
  and Pattern Recognition (CVPR)}.
\bibitem[{Zeng et~al.(2020)Zeng, Xu, Huang, Chen, Tan and Gan}]{Zeng_2020_CVPR}
\bibinfo{author}{Zeng, R.}, \bibinfo{author}{Xu, H.}, \bibinfo{author}{Huang,
  W.}, \bibinfo{author}{Chen, P.}, \bibinfo{author}{Tan, M.},
  \bibinfo{author}{Gan, C.}, \bibinfo{year}{2020}.
\newblock \bibinfo{title}{{Dense Regression Network for Video Grounding}}, in:
  \bibinfo{booktitle}{Proceedings of the IEEE/CVF Conference on Computer Vision
  and Pattern Recognition (CVPR)}.
\bibitem[{Zhang et~al.(2018)Zhang, Hu and Sha}]{Zhang18}
\bibinfo{author}{Zhang, B.}, \bibinfo{author}{Hu, H.}, \bibinfo{author}{Sha,
  F.}, \bibinfo{year}{2018}.
\newblock \bibinfo{title}{Cross-modal and hierarchical modeling of video and
  text}, in: \bibinfo{booktitle}{Proceedings of the European Conference on
  Computer Vision (ECCV)}.
\bibitem[{Zhang et~al.(2019a)Zhang, Dai, Wang, Wang and Davis}]{ZhangDWWD19}
\bibinfo{author}{Zhang, D.}, \bibinfo{author}{Dai, X.}, \bibinfo{author}{Wang,
  X.}, \bibinfo{author}{Wang, Y.}, \bibinfo{author}{Davis, L.S.},
  \bibinfo{year}{2019}a.
\newblock \bibinfo{title}{{MAN:} moment alignment network for natural language
  moment retrieval via iterative graph adjustment}, in:
  \bibinfo{booktitle}{Proceedings of the IEEE/CVF Conference on Computer Vision
  and Pattern Recognition (CVPR)}.
\bibitem[{Zhang et~al.(2020)Zhang, Peng, Fu and Luo}]{zhang2020learning}
\bibinfo{author}{Zhang, S.}, \bibinfo{author}{Peng, H.}, \bibinfo{author}{Fu,
  J.}, \bibinfo{author}{Luo, J.}, \bibinfo{year}{2020}.
\newblock \bibinfo{title}{Learning 2d temporal adjacent networks for moment
  localization with natural language}, in: \bibinfo{booktitle}{Proceedings of
  the AAAI Conference on Artificial Intelligence}, pp.
  \bibinfo{pages}{12870--12877}.
\bibitem[{Zhang et~al.(2019b)Zhang, Su and Luo}]{ZhangSL2019}
\bibinfo{author}{Zhang, S.}, \bibinfo{author}{Su, J.}, \bibinfo{author}{Luo,
  J.}, \bibinfo{year}{2019}b.
\newblock \bibinfo{title}{Exploiting temporal relationships in video moment
  localization with natural language}, in: \bibinfo{booktitle}{ACM MM},
  \bibinfo{publisher}{ACM}.
\bibitem[{Zhang et~al.(2019c)Zhang, Lin, Zhao and Xiao}]{zhang2019cross}
\bibinfo{author}{Zhang, Z.}, \bibinfo{author}{Lin, Z.}, \bibinfo{author}{Zhao,
  Z.}, \bibinfo{author}{Xiao, Z.}, \bibinfo{year}{2019}c.
\newblock \bibinfo{title}{Cross-modal interaction networks for query-based
  moment retrieval in videos}, in: \bibinfo{booktitle}{Proceedings of the 42nd
  International ACM SIGIR Conference on Research and Development in Information
  Retrieval}, pp. \bibinfo{pages}{655--664}.
\bibitem[{{Zhijie Lin, Zhou Zhao, Zhu Zhang, Qi Wang, Huasheng
  Liu}(2020)}]{SCN_2020_AAAI}
\bibinfo{author}{{Zhijie Lin, Zhou Zhao, Zhu Zhang, Qi Wang, Huasheng Liu}},
  \bibinfo{year}{2020}.
\newblock \bibinfo{title}{{Weakly-Supervised Video Moment Retrieval via
  Semantic Completion Network}}, in: \bibinfo{booktitle}{Proceedings of the
  AAAI Conference on Artificial Intelligence}.
\bibitem[{Zhou et~al.(2019)Zhou, Kalantidis, Chen, Corso and
  Rohrbach}]{ZhouKCCR2019}
\bibinfo{author}{Zhou, L.}, \bibinfo{author}{Kalantidis, Y.},
  \bibinfo{author}{Chen, X.}, \bibinfo{author}{Corso, J.J.},
  \bibinfo{author}{Rohrbach, M.}, \bibinfo{year}{2019}.
\newblock \bibinfo{title}{Grounded video description}, in:
  \bibinfo{booktitle}{Proceedings of the IEEE/CVF Conference on Computer Vision
  and Pattern Recognition (CVPR)}.

\end{thebibliography}

\newpage
\appendix
\section{Appendix}

\begin{table*}[!ht]
\caption{
\label{tab:corpus_obj_features}
{\bf  \footnotesize Ablation of different object features for the video corpus retrieval (exhaustive setting).}
\footnotesize  We show average recall for top $K$ retrievals and median retrieval rank (MR, lower is better) averaged over IoU=$\{0.5,0.7\}$ on DiDeMo~\cite{hendricks2017localizing} and Charades-STA~\cite{gao2017tall} datasets for the STAL (TEF) model using different objects features. Model capacity reports the number of parameters (in millions). More details in text.
}
\normalsize
\centering 
\footnotesize
\setlength{\tabcolsep}{4.7pt}
\begin{tabular}{lcc|ccccc|ccccc} 
\toprule 

& 
\multicolumn{2}{c|}{Object Features} &
\multicolumn{5}{c|}{DiDeMo~\cite{hendricks2017localizing} (test)}  & 
\multicolumn{5}{c}{Charades-STA~\cite{gao2017tall} (test)}
\\

&
\multirow{2}{*}{ Glove}  &
\multirow{2}{*}{Region}   &
\multirow{2}{*}{$K$=1}    &
\multirow{2}{*}{$K$=10}   &
\multirow{2}{*}{$K$=100}  &
\multirow{2}{*}{MR $\downarrow$} &
Model &
\multirow{2}{*}{$K$=1}    &
\multirow{2}{*}{$K$=10}   &
\multirow{2}{*}{$K$=100}  &
\multirow{2}{*}{MR $\downarrow$} & 
Model \\

&  &  &
& & & & Capacity &
& & & & Capacity \\
\midrule

a. STAL (TEF) &
\cmark        &
\xmark        &
2.25          &  
10.40         & 
36.46         &
234           & 
12.77         &
{0.31}        &
{2.02}        &
{9.80}        & 
{\bf 2751}    &
10.17.        \\

b. STAL (TEF) &
\xmark     &
\cmark     &
{\bf 2.28} &
10.93      &
{\bf 37.48}&
{\bf 219}  &
16.27      &
\bf{0.39}  &
1.95       &
8.90       &
5173       &
60.42      \\

c. STAL (TEF) & 
\cmark        &
\cmark        &
2.13          &
{\bf 11.00}   &
36.19         &
222           &
16.87         &
0.36          &
\bf{2.06}     &
\bf{9.93}     &
4854          &
60.57         \\

\bottomrule 
\end{tabular}
\end{table*}

We complement our work with the following:

\subsection{Ablation study details} 
\label{sec:ablation-details}

\begin{itemize}
    \item A video summarizing our work is available at \url{http://bit.ly/3nnDrF9}. It shows additional qualitative results generated by our \modelname{} model (\modelnameabbr{})
    
    \item Details about our ablation study (Section~\ref{sec:ablation-details}).
    
    
    
    
    \item  Additional ablation of different object region features (Section \ref{sec:object-region-features}).
    
    \item Results of the Mixture of Embedding Experts~\cite{Miech2018} (MEE) baseline (Section \ref{sec:mee_results}).
    
    
    \item Implementation details and results of the MCN baseline (Section \ref{sec:mcn-details}).
    
    
    
    
\end{itemize}

We clarify the ablation study in Section~\ref{sec:video_corpus} as follows. For moment $\moment^{(\video)}$ with spatiotemporal features $\featurevideo$, 
let $\featurevideo_{\textrm{Clips}}$ be the clip features within the moment (we denote the moment using only clip features as $\moment_{\textrm{Clips}}^{(\video)}$) and let $\featurevideo_{\textrm{Obj}}$ be the detected object features (we denote the moment using only detected object features as $\moment_{\textrm{Obj}}^{(\video)}$). 
For the ``Clips'' ablation, we optimize the squared symmetric Chamfer loss using only the clips features,
\begin{equation}
    \costalignment_{\textrm{Clips}} = \costalignment{\left(\moment_{\textrm{Clips}}^{(\video)}, \languagequery\right)}. 
\end{equation}
For the ``Objects'' ablation, we optimize the squared symmetric Chamfer loss using only the detected objects features,
\begin{equation}
    \costalignment_{\textrm{Objects}} = \costalignment{\left(\moment_{\textrm{Obj}}^{(\video)}, \languagequery\right)}.
\end{equation}
For the ``Clips+obj.\ (early)'' ablation, we optimize the squared symmetric Chamfer loss using the combined set of clip and detected objects features,
\begin{equation}
    \costalignment_{\textrm{Clips+obj.\ (early)}} = \costalignment{\left(\left\{\moment_{\textrm{Clips}}^{(\video)}, \moment_{\textrm{Obj}}^{(\video)}\right\}, \languagequery\right)}.
\end{equation}
For the ``Clips+obj.\ (late-sep)'' ablation, we optimize the sum of two squared symmetric Chamfer losses over clip and detected objects features, which are optimized separately during training and fused together with scalar hyperparameter $\lambda$ during inference (found via cross validation),
\begin{equation}
    \costalignment_{\textrm{Clips+obj.\ (late-sep)}} = \costalignment{\left(\moment_{\textrm{Clips}}^{(\video)}, \languagequery\right)} + 
    \lambda \costalignment{\left(\moment_{\textrm{Obj}}^{(\video)}, \languagequery\right)}.
\end{equation}
For the ``Clips+obj.\ (late-joint)'' ablation, we optimize the sum of two squared symmetric Chamfer losses over clip and detected objects features jointly during training,
\begin{equation}
    \costalignment_{\textrm{Clips+obj.\ (late-joint)}} = \costalignment{\left(\moment_{\textrm{Clips}}^{(\video)}, \languagequery\right)} + 
    \costalignment{\left(\moment_{\textrm{Obj}}^{(\video)}, \languagequery\right)}.
\end{equation}
The latter performs best (\cf, Table~\ref{tab:ablation_fusion}) and is illustrated in Figure~\ref{fig:model-details}.

\begin{table*}[!ht]
\caption{
\label{tab:mee_results}
\textbf{\footnotesize Text to video retrieval results.}
\footnotesize We show Recall@K on MSR-VTT~\cite{Xu2016MSRVTTAL}, DiDeMo~\cite{hendricks2017localizing}, and Charades-STA~\cite{gao2017tall} with RGB features. See text for more details.
}
\centering 
\footnotesize
\scalebox{1.0}{
\begin{tabular}{llcccc} 
\toprule
Dataset & Method & R@1 & R@10 & R@100 & MR $\downarrow$  \\ 
\midrule

\multirow{3}{*}{MSR-VTT~\cite{Xu2016MSRVTTAL}}

& Chance & 0.10 & 0.50 & 1.00 & 500 \\

& Dual Encoding \cite{Dong18} & 7.70 & 22.00 & 31.80 & 32 \\

& MEE \cite{Miech2018} & 11.90 & 34.00 & 48.10 & 11 \\

\midrule

\multirow{3}{*}{DiDeMo~\cite{hendricks2017localizing}}

& Chance & 0.10 & 0.48 & 0.95 & 519 \\

& MEE (trained on MSR-VTT) & 0.10 & 0.40 & 0.94 & 532 \\

& MEE (trained on DiDeMo) & 0.88 & 3.65 & 6.63 & 186 \\

\midrule

\multirow{3}{*}{Charades-STA~\cite{gao2017tall}}

& Chance & 0.08 & 0.38 & 0.75 & 667 \\

& MEE (trained on MSR-VTT) & 0.10 & 0.38 & 0.83 & 664 \\

& MEE (trained on Charades-STA) & 0.48 & 1.78 & 3.08 & 442 \\

\bottomrule
\end{tabular}
}
\end{table*}

\begin{table*}[!ht]
\caption{\label{tab:didemo_original_criteria} 
{\bf \footnotesize Results on DiDeMo (validation and testing set) for single video moment retrieval.} \footnotesize $\star$ corresponds to our implementation, $\circ$ values are taken from the GitHub page. Note that TGN~\cite{Chen18} may not have used the original DiDeMo criteria (they used Rank@$\{1,5\}$ IoU=1). We include their numbers for completeness.}
\centering 
\scalebox{.75}{
\begin{tabular}{llccc|ccc} 
\toprule
 & & \multicolumn{3}{c|}{Validation set} & \multicolumn{3}{c}{Test set} \\
&  & ~Rank@1 & Rank@5 & mIoU~ & ~Rank@1 & Rank@5 & mIoU~ \\ 
\midrule

ACRN~\cite{Liu2018AttentiveMR} & (VGG,Glove) & 
- & - & - &
13.03 & 39.27 & 27.22 \\

TGN~\cite{Chen18} & (VGG,Glove) &
- & - & - &
24.28 & 71.43 & {\bf 38.62} \\

MCN$^\circ$~\cite{hendricks2017localizing} & (VGG,Glove) &
24.42 & 75.40 & 37.39&
23.12 & 73.36 & 35.49 \\

MCN$^\star$~\cite{hendricks2017localizing} & (ResNet,Glove) &
{\bf 26.02} & {\bf 79.52} & {\bf 39.04}~ &
{\bf 25.55} & {\bf 77.19} & 37.82  \\

\bottomrule
\end{tabular}
}
\end{table*}

\subsection{Object region features}
\label{sec:object-region-features}

Here we investigate the pros and cons of different object features in our STAL (TEF) model. Object features used in the main experiments (Section~\ref{sec:experiments}) are based on the Glove word embedding of class names of the top 10 detected objects in each video clip. The intuition is that such representation is a compact abstraction of the object class depicted in the clips and is directly comparable to the glove embedding of the natural language query. Here we investigate replacing this object representation with the region features extracted from the $pool5$ layer of the Faster R-CNN architecture, which directly represent the object appearance. 
Finally, we also investigate concatenating the $pool5$ region features together with the Glove embedding feature into a single representation. Results are shown in Table~\ref{tab:corpus_obj_features} on the ``Exhaustive Search setup''. The Glove word embedding only, as reported in Table~\ref{tab:corpus_quantitative}, is shown in the first row (a.). $Pool5$ region features only are shown in the second row (b.) and their combination in the third row (c.). Next, we evaluate the pros/cons of the different object representations in terms of retrieval accuracy, resulting model size and memory requirements for storage and training.

\paragraph{Retrieval accuracy.} In terms of retrieval accuracy, the object region features provide only marginal improvements in recall@K accuracy for the DiDeMo~\cite{hendricks2017localizing} and Charades-STA~\cite{gao2017tall} datasets. We believe that the Glove object features already provide the necessary information for bridging the visual and textual information and provide informative hints for the retrieval task. Moreover, by construction, the language features and Glove object embeddings live in the same feature space. This can potentially alleviate the semantic alignment task performed by our architecture. In contrast, the region feature space is different. Therefore when using the region features, our network must also learn a good mapping between the region feature space and the language feature space before attempting the alignment of the two modalities. Nonetheless, we acknowledge that the region features might help to discriminate between two different object instances (of the same class) appearing in the same scene. Learning a more complex fusion of the Glove and region features is an interesting direction for future work.

\paragraph{Model size (capacity).} The higher dimensionality of the feature representation also induces an increase in the model size. As shown in Table~\ref{tab:corpus_obj_features}, when replacing the Glove features with the region features for the DiDeMo dataset, the number of model parameters increases by $27.41$\% (from $12.77$ million to $16.27$ million). The model size is even larger when the two features are concatenated together. For Charades-STA, the hyperparameters search resulted in a much larger network necessary to achieve the same retrieval performance. There is a $6\times$ increase in model capacity between setup (a.) and (b.).

\paragraph{Memory and storage requirements.} The different object features have different storage requirements, which affect feature loading time and memory needs during training. Using the Glove embedding of the detected class lets us efficiently store the information as an integer value (class index), while the encoding operation for the class name is performed only once at the beginning of the training process.
Conversely, the object regions must be stored, and therefore occupy $4096$ times more space ($16$ bits for a single integer vs. $2048 \times 32$ bits for floats). At training time, when each batch is created and moved from RAM memory to GPU memory, region features require $\sim6.8$ more GPU memory, as the Glove embedding feature dimension is only $300$ (vs. $2048$ for the region features). 

\begin{table*}[!ht]
\caption{
\label{tab:single_video_retrieval}
{\bf \footnotesize Results on  Charades-STA~\cite{gao2017tall} for single video moment retrieval.}
 \footnotesize $\star$ corresponds to our implementation.
}
\centering 
\footnotesize
\scalebox{1.0}{%
\begin{tabular}{ll  cc | cc } 
\toprule 

& &
\multicolumn{4}{c}{Test set} 
\\

& & 
\multicolumn{2}{c|}{IoU=0.5} &
\multicolumn{2}{c}{IoU=0.7} 
\\

& & ~~R@1~ & ~R@5~~ &
~~R@1~ & ~R@5~~ \\ 
\midrule

CTRL~\cite{gao2017tall} & (C3D,Skip-thought) & 
 21.42 & 59.11 &
 7.15 & 26.91 \\ 

\cite{Xu19} & (C3D,word2vec) & 
 35.60 & 79.40 &
 15.80 & 45.40 \\

CBP~\cite{WangXWQLTG2016} & (C3D,word2vec) & 
 36.80 & 70.94 &
 18.87 & 50.19 \\

MCN$^\star$~\cite{hendricks2017localizing} & (ResNet,word2vec) & 
 {\bf 44.48} & {\bf 83.63} &
 {\bf 24.65} & {\bf 56.92} \\

\bottomrule 
\end{tabular}
}
\end{table*}

\subsection{Mixture of embedding experts for Text-To-Video retrieval} \label{sec:mee_results}

In this section, we provide more details about the Mixture of Embedding Experts baseline (MEE) used in Section~\ref{sec:video_corpus_twostage} ~\cite{Miech2018}. This model belongs to the family of methods of video clip retrieval with natural language described in Section~\ref{sec:related_work}.
In a nutshell, given a natural language query, MEE retrieves the most similar trimmed video clip that aligns with the given query.
This approach, by itself, falls short of addressing our proposed task of retrieving relevant moments from a large corpus of untrimmed, unsegmented videos. Thus, we paired MEE with a model for localizing moments in a single video in a two-stage fashion to fulfill the requirement of our task.

We chose MEE over other methods as it performs well on LSMDC benchmark and outperforms other recent methods~\cite{Dong18} on MSR-VTT~\cite{Xu2016MSRVTTAL}. 
Table~\ref{tab:mee_results} shows the quantitative comparison in the text to video retrieval task on the MSR-VTT corpus (\cf, ``MSR-VTT''). 
We used the publicly available implementation of MEE to retrieve videos from the corpus and turned off flow, face and audio features in MEE for fair comparison.

For our two-stage retrieval baseline of moments from a video corpus for a natural language query, we tested MEE models pre-trained on MSR-VTT and the corresponding dataset. Table~\ref{tab:mee_results} summarizes the results of these experiments (\cf, ``DiDeMo, Charades-STA''). We observed that MEE ~pre-trained on MSR-VTT performed near chance; while re-training MEE on the target datasets performed the best. Thus, we used the latter setup in the rest of the experiments in our main submission.

\subsection{Additional details of MCN baseline}
\label{sec:mcn-details}

For fair and extensive experimentation, we use our own implementation of ~MCN~\cite{hendricks2017localizing} in ~Pytorch~\cite{pytorch} and NumPy~\cite{numpy}.
In this section, we disclose the results of our implementation on the single video retrieval task proposed by~\cite{hendricks2017localizing,gao2017tall}.
Table~\ref{tab:didemo_original_criteria} presents the results in the validation and test set of DiDeMo dataset using the original metric proposed by~\cite{hendricks2017localizing}.
We report (carbon copy) the original MCN~\cite{hendricks2017localizing} performance (MCN$^\circ$~\cite{hendricks2017localizing}) as well as other publicly available models, and compare against our own implementation (MCN$^\star$~\cite{hendricks2017localizing}).
We observe that our implementation is more accurate over all metrics in the validation set, and all except for mIoU in the test set.
Table~\ref{tab:single_video_retrieval} shows the performance of our implementation in Charades-STA dataset using the standard evaluation protocol from~\cite{gao2017tall}.
Here we also compare with publicly available methods, and again our implementation achieves better accuracy than existing alternatives.
Note that it is not possible to evaluate the public models of MCN~\cite{hendricks2017localizing} on Charades-STA as pre-computed features are not available on the dataset. 
In sum, we conclude that MCN is a strong baseline for this task, and also the only known approach for efficient single video moment retrieval via approximate nearest neighbours.

\end{document}